\documentclass[11pt]{article}

\usepackage[final]{acl}

\usepackage{times}
\usepackage{latexsym}
\usepackage[T1]{fontenc}
\usepackage[utf8]{inputenc}
\usepackage{microtype}
\usepackage{amsmath}
\usepackage{amssymb}
\usepackage{booktabs}
\usepackage{graphicx}
\usepackage{url}
\usepackage{hyperref}

\title{Shadow-Loom: Causal Reasoning over Graphical World Models of Narratives}

\author{David Wilmot \\
        david.wilmot@gmail.com \\
        \texttt{\href{https://github.com/dwlmt/shadow-loom}{github.com/davidrwilmot/shadow-loom}} \\
        Released under AGPL-3.0-or-later + Non-Commercial Clause}

\begin{document}
\maketitle

\begin{abstract}
Stories hold a reader's attention because they have causes,
secrets, and consequences. \textbf{Shadow-Loom} is an experimental
open-source framework that turns a narrative into a versioned
\emph{graphical world model} and lets two engines act
on it: a \emph{causal physics} grounded in Pearl's ladder of
causation \citep{pearl2009causality,bareinboim2022pearl} and the
recently proposed counterfactual calculus over Ancestral
Multi-World Networks \citep{correa2025amwn}; and a
\emph{narrative physics} that scores the same graph against four
structural reader-states --- mystery, dramatic irony, suspense,
and surprise --- in the tradition of Sternberg's curiosity /
suspense / surprise triad \citep{sternberg1992telling}, with
suspense formalised in the structural-affect line of
\citet{brewer1982stories}, \citet{comisky1982suspense} and
\citet{cheong2015suspenser}. Large language models are used
only at the boundary: extraction, rendering, and audit;
identification, intervention, and counterfactual reasoning are
carried out in typed code over the graph. The system is offered
as a research artefact rather than as a benchmarked NLP model;
code, fixtures, and pipeline are released open source.
\end{abstract}

\section{Introduction}
\label{sec:intro}

Most current LLM-driven story systems treat a narrative as a
sequence-prediction problem and absorb the causal, temporal, and
epistemic structure of the story into the weights of a single
model. Recent work shows that such models are competent at
\emph{reading} causal relations from text
\citep{kiciman2024causal} but markedly weaker at \emph{acting}
on the implicit beliefs of characters they have just inferred
\citep{gu2026simpletom,kim2023fantom}, and similarly weak when
asked to reason hypothetically about other agents' strategies
without external scaffolding \citep{cross2024hyp}; in
unconditioned generation they tend to collapse onto a small set
of shared plot tropes \citep{xu2025echoes,tian2024human}. The
complementary direction explored here keeps an explicit, typed
graph of the storyworld
\citep{schank1977scripts,mani2012computational,elson2012sig} and
performs identification, intervention, and counterfactual
reasoning over it symbolically
\citep{pearl2009causality,halpern2016actual,correa2025amwn},
using the LLM only as a constrained renderer
\citep{riedlyoung2010narrative,yao2019planwrite,
goldfarbtarrant2020aristotelian,yang2023doc}. Concretely, a
query such as ``what if Macbeth refuses to murder Duncan?'' becomes
a rung-2 (Intervention) do-operation on a typed event graph extracted from the
play (App.~\ref{app:ex-amwn}) rather than a free-form prompt to
a language model; ``does Poirot know the conspirators are still
lovers?'' becomes a finite belief-set membership test
(App.~\ref{app:ex-narrative}) rather than a single forward pass.

\paragraph{Motivation 1: counterfactuals are where LLMs tend to
struggle.} Pearl's hierarchy is provably non-collapsible. Rung-3
quantities (Counterfactuals) are not, in general, identifiable from rung-2
(Intervention) data, let alone from rung-1 (Observation) data
\citep{bareinboim2022pearl}.
Empirically, LLMs reproduce surface causal patterns from their
training corpus \citep{kiciman2024causal} but tend to degrade
sharply on multi-hop belief-conditioned action prediction
\citep{kim2023fantom,gu2026simpletom} and on story-internal
counterfactual consistency \citep{xu2025echoes}. The questions
that ordinarily matter to a reader of a story --- ``what if
Macbeth had refused?'', ``what if Roy had told Mary the truth?''
--- live on the rung where the modern foundations (AMWN +
ctf-calculus, \citealp{correa2025amwn}) are designed to operate,
and where end-to-end neural generators are not. A typed graph
plus an explicit do-operator and abduction step turns each such
question into a finite sequence of well-defined operations
\citep{pearl2009causality,halpern2005causes,halpern2016actual}
rather than a free-form imagination task.

\paragraph{Motivation 2: coding agents do better, arguably,
because they compile and test.} A familiar pattern in agent
design is that coding agents tend to outperform open-domain
agents at fixed model capacity \citep{jimenez2024swebench}.
Their outputs can be compiled, executed against a test suite,
and rejected when they fail. The
lever, on this reading, is the \emph{typed substrate underneath
the natural-language reasoning}, not the model itself.
Shadow-Loom borrows that lever for narrative. A continuation is
rejected if any of three checks fails: (a) its causal
propagation (Eq.~\ref{eq:impact}) violates inertia or
affordance gating; (b) its counterfactual branch fails an
AMWN-style consistency, independence, or exclusion check
\citep{correa2025amwn}; or (c) its re-extraction does not match
the brief that licensed it (the audit pass, in the
LLM-as-judge tradition of
\citealp{bai2022constitutional,madaan2023selfrefine,gu2024judge}).
The graph plays something like the role of the compiler; the
auditor plays something like the role of the test suite.

The framework combines four ideas which, so far as I can tell,
have not previously been wired together in a single open-source
pipeline:

\begin{enumerate}
\item A \textbf{Pydantic-typed world graph} (\texttt{WorldStateV1})
      with entities, events, beliefs, locations, channels, and three
      kinds of edge (causal, social, spatial), every node and edge
      tagged with both a chronological \emph{fabula} time and a
      narrative \emph{syuzhet} index \citep{shklovsky1965art,
      genette1980narrative}.
\item A \textbf{causal physics engine} --- a graphical world model
      of cause and effect --- that implements Pearl's three
      rungs --- observation, do-intervention, and abductive
      counterfactual --- with a propagation rule gated by an
      explicit ``Impact $>$ Inertia'' threshold and spatial
      affordance checks.
\item A \textbf{narrative physics layer} --- a graphical world
      model of reader affect over the same graph --- that scores it
      against closed-form information-state functionals for mystery,
      dramatic irony, suspense, and surprise. Suspense uses the
      hope/fear (here \emph{hope/threat}) anticipation pair from the
      structural-affect tradition
      \citep{brewer1982stories,comisky1982suspense,ortony1988occ,
      zillmann1996suspense} as operationalised by
      \citet{cheong2015suspenser}, combined as a
      \emph{balance$\,\times\,$stakes} product over the unrevealed
      ledger; this is a deliberate departure from the
      information-theoretic uncertainty-reduction formulation of
      \citet{wilmot2020suspense}, whose reader-vs-future-state
      epistemic-asymmetry shape we instead borrow for the dramatic
      irony scorer in \S\ref{app:narrative}. Surprise is binary KL
      divergence between a leave-one-out corpus-marginal prior,
      evidence-updated by
      revealed causal edges, and the entity's reconstructed final
      state.
\item A \textbf{constrained-LLM renderer + recursive auditor} loop
      \citep{gu2024judge} that converts the simulation result into
      a typed creative brief, generates prose under that brief, and
      then re-extracts the prose to verify no ``miracle steps''
      were introduced.
\end{enumerate}

The system is named for the dual-graph metaphor: a \emph{loom} of
factual edges runs in parallel with one or more \emph{shadow}
counterfactual branches, in the sense of the AMWN construction of
\citet{correa2025amwn}. Section~\ref{sec:pipeline} sketches the
pipeline; Section~\ref{sec:novelty} states what is novel and what
is borrowed; Section~\ref{sec:rationale} gives the design
rationale; Section~\ref{sec:relevance} discusses relevance to NLP,
reasoning, and social science; Section~\ref{sec:limits} states the
limitations honestly. Equations and per-stage definitions are in
App.~\ref{app:defs}; an extended narrative walkthrough of the
whole loop on a single fixture is in App.~\ref{app:walkthrough};
worked examples drawn from the twenty bundled plot fixtures
follow in App.~\ref{app:examples}; and a separate appendix on the
authoring user interface, with screenshots and step-by-step
example sessions, is in App.~\ref{app:ui}.

\paragraph{How to read this paper.} The main text is intentionally
short. It states the design, situates it in the literature, and
records what is novel as a combination. Readers who want to
understand the symbolic machinery in detail --- the schema, the
ego-graph extraction, the do-operator, the abductive
counterfactual, the four affective scorers, the brief-and-audit
loop --- should treat App.~\ref{app:defs} as the authoritative
specification. App.~\ref{app:walkthrough} is a long-form
description of the same pipeline expressed as a story:
\emph{Macbeth} is fed in as raw prose and followed step-by-step
through ingestion, intervention, propagation, scoring, rendering,
and audit, with every intermediate object shown. The eight
example sub-appendices in App.~\ref{app:examples} then show how
each individual stage manifests on a different bundled fixture.
App.~\ref{app:ui} is for readers who want to understand or use the
authoring workspace; it cross-links every UI surface to the
pipeline stage it exposes and includes example sessions on real
fixtures.

\section{Pipeline overview}
\label{sec:pipeline}

\begin{figure}[h]
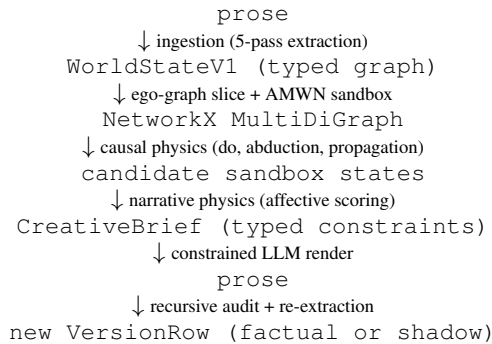

\centering
\small
\begin{tabular}{c}
\texttt{prose}\\
$\downarrow$ \scriptsize ingestion (5-pass extraction)\\
\texttt{WorldStateV1 (typed graph)}\\
$\downarrow$ \scriptsize ego-graph slice + AMWN sandbox\\
\texttt{NetworkX MultiDiGraph}\\
$\downarrow$ \scriptsize causal physics (do, abduction, propagation)\\
\texttt{candidate sandbox states}\\
$\downarrow$ \scriptsize narrative physics (affective scoring)\\
\texttt{CreativeBrief (typed constraints)}\\
$\downarrow$ \scriptsize constrained LLM render\\
\texttt{prose}\\
$\downarrow$ \scriptsize recursive audit + re-extraction\\
\texttt{new VersionRow (factual or shadow)}\\
\end{tabular}
\caption{The Shadow-Loom loop. The LLM is invoked only at the top
(ingestion), the bottom (rendering), and the audit; everything in
between is symbolic computation over the graph. See the appendix
for definitions of each stage.}
\label{fig:pipeline}
\end{figure}

The pipeline is a five-phase loop over a versioned world model.
Every version is ancestor-linked, so a counterfactual fork is
literally a sibling row in a directed acyclic version tree, not a
mutation of canon. Figure~\ref{fig:pipeline} sketches the loop.

The five phases correspond to the five appendix subsections
(\ref{app:ingestion}--\ref{app:audit}):
\textbf{(1) Ingestion} runs five LLM agents to extract a typed
\texttt{GlobalRegister}, then three concurrent specialist agents
(physics, social, consequences) per chunk to extract events,
relationships, channels, and entity-state deltas, with a
programmatic validation and auto-repair loop.
\textbf{(2) Causal physics} forks an Ancestral Multi-World Network
sandbox, applies a do-operator for interventions, runs abduction for
counterfactuals, and propagates effects through a topological sort
of the causal sub-graph subject to inertia and affordance gating.
\textbf{(3) Narrative physics / affective calculus} scores each
candidate sandbox against four structural information-state
functionals and six trait-trajectory emotional functionals.
\textbf{(4) Generation} packages the highest-scoring candidate as a
\texttt{CreativeBrief} of typed \texttt{ConstraintBlock}s and hands
\emph{only} the brief to a creative LLM.
\textbf{(5) Audit and feedback} runs an LLM-as-judge over the prose,
detects ``miracle steps'' (state changes with no licensing edge in
the brief), and either accepts the version or re-runs generation
with the auditor's feedback appended, up to a configurable retry
budget.

The entire loop is exposed both through a NiceGUI authoring
workspace (nine cross-linked tabs over a versioned project DAG;
App.~\ref{app:ui}) and through a Model-Context-Protocol server
with roughly forty tools, so a downstream agent can drive
shadow-branch creation, scoring, and promotion programmatically.

\section{Theoretical lineage and computational counterparts}
\label{sec:lineage}

The system has been built theory-first: every named field in
\texttt{WorldStateV1} corresponds to a specific narratological,
cognitive, or causal-inference construct, and every field has a
prior computational instantiation that the design either
generalises or borrows from. The pairings the design depends on
are summarised below; an extended mapping lives in the project's
academic-foundations document.

\paragraph{Fabula vs syuzhet
\citep{shklovsky1965art,tomashevsky1965thematics,
genette1980narrative,bordwell1985narration,bal2009narratology}.}
Two integer indices on every \texttt{EventNode}: chronological
$f(e)$ and presentation $s(e)$. The four affective scorers reduce
to set operations between the two projections (App.~\ref{app:narrative}).

\paragraph{Possible-worlds narratology
\citep{ryan1991possible,ryan2001virtual,dolezel1998heterocosmica,
pavel1986fictional}, modal-logical roots
\citep{lewis1973causation}, and counterfactual graphical models
\citep{balke1994counterfactual,shpitser2008complete,
correa2021nested,correa2025amwn}.} Every node and edge carries
$\texttt{world\_id} \in \{\text{factual}, \text{shadow}\}$;
counterfactual queries fork a sibling \texttt{VersionRow}.
The AMWN tag and three-rule mental model (consistency,
independence, exclusion) come from \citet{correa2025amwn}; I
adopt the naming and architecture even though the typed narrative
graph used here is not a rigorous SCM. AMWN supersedes the Twin-Network
construction \citep{balke1994counterfactual} and the multi-network
machinery of \citet{shpitser2008complete}, and is the
counterfactual-completeness counterpart to do-calculus
\citep{pearl1995causal,pearl2009causality,bareinboim2022pearl}.

\paragraph{Greimas \citep{greimas1983structural}, Propp
\citep{propp1968morphology}, Bremond \citep{bremond1973logique},
and the symbolic story-grammar tradition
\citep{rumelhart1975schema,mandlerjohnson1977,kintsch1978macro,
schank1977scripts,lehnert1981plotunits,meehan1977talespin,
perez2001mexica,bringsjord2000brutus,chambers2008narrative}.}
\texttt{GlobalTrait} (\texttt{WORLD\_}) realises Greimas's ``Power''
actant; \texttt{RelationshipEdge.power\_dynamic} mirrors the
sender / receiver / helper / opponent topology;
\texttt{EventNode.event\_type} $\in \{\text{choice, outcome,
revelation, utterance}\}$ is a coarse generalisation of Propp's
functions,
closer in spirit to Bremond's triad. The plot-unit and script
traditions prefigure both the typed-event graph
\citep{schank1977scripts,lehnert1981plotunits} and the
goal-directed planner / generator pipeline
\citep{meehan1977talespin,perez2001mexica}; the unsupervised
narrative event chains of \citet{chambers2008narrative} are a direct
precursor to the typed \texttt{causal\_topology} maintained at
runtime.

\paragraph{Cognitive narratology
\citep{herman2002story,fludernik1996natural,zwaan1998situation,
gerrig1993narrative} and constructionist comprehension
\citep{trabasso1985causal,graesser1994constructing,kintsch1978macro}.}
The five fields of the situation-model (time, space, protagonist,
causation, intention) are nearly the field set on \texttt{EventNode};
that readers materialise causal links explicitly during
comprehension is the empirical justification for materialising them
explicitly in storage.

\paragraph{Theory of mind
\citep{premack1978tom,wimmer1983false,baroncohen1985autism,
zunshine2006theory,baker2009inverseplanning,chandra2024inverse}, its modern LLM
benchmarks \citep{kim2023fantom,gu2026simpletom}, and ToM-scaffolded
LLM agents \citep{cross2024hyp}.}
\texttt{Belief\{value, confidence, inertia,
established\_at\_fabula, acquired\_via\_event\_id,
acquired\_via\_channel\_id\}} is a discrete approximation of
recursive ToM with provenance. The benchmarks above motivate the
auditor's separate behaviour-given-belief consistency check.

\paragraph{Sternberg's curiosity / suspense / surprise triad
\citep{sternberg1978expositional,sternberg1992telling,
brewer1982stories,baroni2007tension} and its computational realisations
\citep{cheong2015suspenser,baeyoung2008surprise,
elyfrankel2015suspense,wilmot2020suspense,wilmot2021emnlp,
wilmot2022phd}.} The four structural scorers (\textsc{Mys},
\textsc{Iro}, \textsc{Susp}, \textsc{Sur}) extend the triad with
dramatic irony \citep{booth1974irony,gerrig1993narrative}. The
suspense scorer follows the hope/fear anticipation pair of the
structural-affect and appraisal traditions
\citep{brewer1982stories,comisky1982suspense,ortony1988occ,
zillmann1996suspense,gerrig1989suspense} as operationalised in narrative planning by
\citet{cheong2015suspenser}, combining the two sides as a
\emph{balance$\,\times\,$stakes} product that peaks under genuine
outcome uncertainty rather than under one-sided dominance. This is
a deliberate departure from \citet{wilmot2020suspense}, whose
suspense is information-theoretic uncertainty reduction over neural
state representations and whose decision-theoretic counterpart is
\citet{elyfrankel2015suspense}; that reader-vs-future-state
epistemic-asymmetry shape (the reader knows things the focal future
does not yet contain) is structurally closer to dramatic irony
(reader knows things the characters do not) than to a hope/threat
ledger, and motivates the irony scorer in
App.~\ref{app:narrative} rather than the suspense scorer. Surprise
follows \citet{ittibaldi2009surprise} with a geometric prior update
over a leave-one-out corpus marginal; emotional arc analytics owe
to \citet{reagan2016arcs}.

\paragraph{Information-theoretic and epistemic substrate
\citep{shannon1948mathematical,fagin1995reasoning,
alchourron1985agm}.} \texttt{Channel} carries
\texttt{intelligibility} $\in [0,1]$ per participant
(generalising the legacy encryption boolean) and an
\texttt{eavesdropped\_by} relation derived from the
\texttt{intelligibility\_threshold}; belief revision uses an
AGM-style \texttt{beliefs\_added / beliefs\_invalidated} delta.

\paragraph{Causal inference
\citep{pearl1995causal,pearl2009causality,pearl2018book,
spirtes2000causation,bareinboim2022pearl} and actual causality
\citep{halpern2005causes,halpern2016actual,woodward2003making}.}
Pearl's three rungs --- \textbf{rung~1 Correlation} (observation),
\textbf{rung~2 Interventions} ($\mathrm{do}(\cdot)$), and
\textbf{rung~3 Counterfactuals} (abduction~+~do) --- map directly
to the query taxonomy used here
(\texttt{ObservationQuery}, \texttt{InterventionQuery},
\texttt{CounterfactualQuery}); \texttt{mechanism} and
\texttt{causal\_force} on edges are motivated by Halpern's
``actual cause''; ego-graph slicing is a heuristic Markov-blanket
cut \citep{vermapearl1988dsep}. Recent benchmarking of LLMs as
causal reasoners \citep{kiciman2024causal} supports the hybrid
stance taken here: LLMs propose edges (\texttt{extract\_graph.py});
typed code identifies and propagates them
(\texttt{causal\_physics.py}).

\paragraph{Plan-and-render era of neural story generation
\citep{riedlyoung2010narrative,jhalayoung2010cinematic,
martin2018event,yao2019planwrite,goldfarbtarrant2020aristotelian,
rashkin2020plotmachines,yang2022re3,yang2023doc} and its
benchmarks \citep{mostafazadeh2016cloze,fan2018writingprompts,
akoury2020storium,tian2024human,xu2025echoes}.} The
\texttt{CreativeBrief} (Step 9) plus channel-aware draft
(Step 10) is a typed, causal-graph-conditioned variant of the same
plan-then-render pattern. The brief enumerates per-effect typed
\texttt{ConstraintBlock}s; the auditor (Step 11) is the rescore /
revise step \citep{goldfarbtarrant2020aristotelian,yang2022re3}
specialised to narrative-causal rather than stylistic criteria,
in the LLM-as-judge tradition \citep{zheng2023judge,
bai2022constitutional,madaan2023selfrefine,gu2024judge,
qi2023socratic}, with structured prompting and self-improvement
techniques \citep{wei2022cot,zelikman2022star} used inside the
renderer when free-form reasoning is required.

\paragraph{Reading as simulation
\citep{maroatley2008fiction,oatley2016fiction,oatley1995taxonomy,
hogan2003mind}.} The reason a tight causal/affective model
matters more than surface fluency: fiction is a cognitive
simulator, and what is being simulated is a typed world.

\paragraph{Game engines, interactive narrative, and world models
for AI \citep{mateas2003facade,riedl2013interactive,ware2014glaive,
kreminski2020wereyaknow,park2023generative,ha2018world,lecun2022jepa,
hao2023rap,wang2023voyager,smelik2014procedural,
summerville2018pcgml}.} Architecturally, Shadow-Loom is closer to
a game engine than to a sequence model: a typed simulation step
(causal physics) is wrapped by a director / drama-manager
(narrative physics) and exposed to a stochastic policy (the LLM
renderer) only through a constrained action interface, in the
lineage of Mimesis / Fa\c{c}ade-style interactive drama
\citep{mateas2003facade,riedl2013interactive} and narrative
planners such as CPOCL/Glaive \citep{ware2014glaive}. The
generative-agent line of work \citep{park2023generative} and
the ``world model'' line in deep RL
\citep{ha2018world,lecun2022jepa,hao2023rap} make a similar
wager from the opposite direction: that an explicit, manipulable
latent world is what makes long-horizon coherent behaviour
tractable. Shadow-Loom's shadow branches play the role those
world models play --- a sandbox in which counterfactual
rollouts can be evaluated before being committed --- but with a
typed, inspectable schema in place of an opaque latent state.
The bundled example worlds and authoring tools relate
Shadow-Loom to the broader procedural-content-generation
tradition
\citep{smelik2014procedural,summerville2018pcgml,kreminski2020wereyaknow}.

\section{What is novel, what is borrowed}
\label{sec:novelty}

\paragraph{Borrowed.} Every ingredient in
Section~\ref{sec:lineage} is borrowed: Pearl's ladder
\citep{pearl2009causality,bareinboim2022pearl}; the AMWN +
ctf-calculus framework \citep{correa2025amwn}; actual-causality
semantics \citep{halpern2016actual}; the structural-affect /
hope-fear anticipation pair underlying the suspense scorer
\citep{brewer1982stories,comisky1982suspense,ortony1988occ,
zillmann1996suspense,cheong2015suspenser}; the
uncertainty-reduction view of suspense (re-used here as the
epistemic-asymmetry shape of the irony scorer) of
\citet{wilmot2020suspense}; the Bayesian-surprise formalism of
\citet{ittibaldi2009surprise}; the LLM-as-judge audit pattern
\citep{bai2022constitutional,madaan2023selfrefine,gu2024judge};
the plan-and-render line in neural story generation
\citep{riedlyoung2010narrative,yao2019planwrite,
goldfarbtarrant2020aristotelian,yang2023doc,yang2022re3,
rashkin2020plotmachines}; the typed-event / story-grammar tradition
\citep{schank1977scripts,lehnert1981plotunits,mani2012computational,
elson2012sig}; and the narratological foundations
(\citealp{shklovsky1965art,genette1980narrative,sternberg1992telling,
ryan1991possible,herman2002story,zwaan1998situation}).

\paragraph{Novel as a combination.} I am not aware of another
open-source pipeline that, taken together: (a) runs all three
Pearl rungs over a typed narrative graph; (b) tags every node
and edge with an AMWN \texttt{world\_id}
\citep{correa2025amwn} so that counterfactuals are first-class
persisted objects; (c) couples that machinery to closed-form
scorers for the four classical reader-states of narrative
cognition; (d) uses the resulting score as a loss function over
candidate interventions, with the LLM constrained to render the
winner; and (e) closes the loop by re-extracting the rendered
prose and comparing it against the brief. Each ingredient is
borrowed. The composition is the modest contribution: a typed
two-engine substrate underneath a constrained renderer.
Architecturally the closest analogue is something like a game
engine for narrative: a typed simulation step gated by a
drama-manager-like scorer
\citep{mateas2003facade,riedl2013interactive,ware2014glaive},
driving a constrained stochastic actor in the spirit of recent
generative-agent and world-model work
\citep{park2023generative,ha2018world,hao2023rap}, but with the
latent state replaced by an explicit causal-graphical schema.

\paragraph{Specific small constructions.} The
``Impact $>$ Inertia'' propagation rule (Eq.~\ref{eq:impact}) and
the geometric prior update inside the surprise scorer
(Eq.~\ref{eq:surprise}) are introduced here as deliberately simple
choices that keep both physics monotonic in useful ways
(App.~\ref{app:narrative}). The two-axis fabula/syuzhet sampling
of the affective scorers (App.~\ref{app:narrative}) is, to my
knowledge, also new.

\section{Rationale for the design choices}
\label{sec:rationale}

\paragraph{Why a typed graph rather than free text + RAG?}
Recent stress tests show that LLMs can answer ``what does X
believe?'' but fail to behave consistently with that belief in
downstream action prediction \citep{gu2026simpletom,kim2023fantom}.
Long-context reading itself is unreliable in the same regime
\citep{liu2024lost}, which compounds the problem when the
``world'' has to be reconstructed from a growing prose buffer.
Belief and causal structure that are recoverable from prose are
not, on present evidence, reliably \emph{operated on} inside the
prose-only loop. A typed graph makes belief, channel
intelligibility, and causal mechanism inspectable and
intervenable.

\paragraph{Why both a causal and a narrative physics?}
The causal physics decides \emph{what is allowed to happen} given
the graph; the narrative physics decides \emph{which of the allowed
things would be most readable}. Decoupling licensing from
desirability lets the system score and rank candidate
interventions before any text is generated, and lets the same
engine serve both ``what-if'' analytical queries (rung-2 Intervention
and rung-3 Counterfactual) and
high-level affective directives (``maximise dramatic irony for
character X'').

\paragraph{Why constrain the LLM rather than fine-tune one?}
The renderer's job is reduced to dialogue, description, and
pacing within a mathematical envelope. Whatever the model's
training distribution, the brief enumerates the causal edges,
beliefs to preserve, channels to keep hidden, and trait shifts
to honour. Errors that survive into prose are then catchable by
the audit pass because they correspond to specific, testable
claims --- the ``compile and test'' analogue from
Motivation~2.

\paragraph{Why AMWN tagging at the persistence layer?}
Treating a counterfactual as a sibling \texttt{VersionRow} with
\texttt{world\_id="shadow"} makes ``what-if'' branches first-class,
diffable, promotable, and auditable; it is the persistence-layer
analogue of the AMWN graphical construction
\citep{correa2025amwn}. A user can compare a shadow against canon,
keep both, or promote.

\section{Why this might matter beyond fiction}
\label{sec:relevance}

\paragraph{NLP / reasoning.} Narrative is, for these purposes, a
reasonably tractable test bed for counterfactual and
theory-of-mind reasoning, because the ground truth (what happens,
who knows what) is finite and authored. A typed-graph + symbolic
physics + constrained-renderer architecture is one concrete
instantiation of the neuro-symbolic stance for controllable
generation; the audit pass produces a falsifiable record of
where the LLM departed from the brief. Recent ToM benchmarks
\citep{gu2026simpletom,kim2023fantom} and ToM-scaffolded LLM
agents \citep{cross2024hyp} ask exactly
the kind of behaviour-given-belief questions that an explicit
\texttt{Belief} field with provenance is designed to support.

\paragraph{Computational social science and digital humanities.}
Many socio-historical and humanistic questions are counterfactual
in form (``what if policy $X$ had not been adopted?'', ``how
would this novel read if the protagonist had spoken sooner?''). A
graphical substrate that lets a researcher ingest a narrative
source, fork a shadow branch under an explicit intervention, and
inspect the propagated trait and relationship deltas is a small
step toward auditable narrative simulation that can sit alongside
established computational-social-science
\citep{lazer2009css} and digital-humanities tooling. I make no
claim of empirical validity for any specific simulation; the
contribution is the substrate, not its calibration.

\paragraph{Author tooling.} The same machinery surfaces in a
NiceGUI workspace and a Model-Context-Protocol server with around
forty tools, exposing the graph, the affective scorers, and the
version DAG to either human authors or LLM agents. The workspace
is described component-by-component in App.~\ref{app:ui}.

\paragraph{An invitation to other research communities.}
The substrate is deliberately domain-agnostic at the graph level:
entities, events, beliefs, channels, and counterfactual branches
are not specific to fiction. Computational social scientists
\citep{lazer2009css} routinely treat narrative sources --- court
records, parliamentary debates, oral histories, news streams ---
as data, and digital-humanities scholarship has long sought
structured, sharable representations of literary works that can
support both close and distant reading. Shadow-Loom may be read
as one such representation: an open, typed, version-controlled
graph with an explicit counterfactual sandbox and an audit trail
back to the source text. Researchers in those communities, as
well as in narratology and cognitive science, are warmly invited
to fork the schema, swap in their own ingestion prompts and
affective scorers, and report what they find. The pipeline ships
with twenty worked example worlds drawn from canonical
literature and film, which may serve as a common reference set
for cross-group experimentation.

\section{Limitations and future evaluation}
\label{sec:limits}

Shadow-Loom is an experimental research artefact rather than a
benchmarked NLP system, and I want to be explicit about what that
does and does not imply. No headline numbers are reported against
an external dataset: the scores produced by the narrative physics
are design choices grounded in narratological theory rather than
validated reader-reaction models, and the causal physics borrows
the AMWN naming and three-rule mental model of
\citet{correa2025amwn} without implementing their identification
algorithm literally --- the graph is heavily typed for narrative
use rather than for rigorous structural causal model
identification. The ingestion pipeline depends on LLM extraction,
and although a programmatic validation and auto-repair loop
catches the more obvious failure modes, extraction errors can
still propagate. The current unit-test suite covers the symbolic
layers but does not amount to external evaluation.

Formal evaluation is therefore deferred to future work, along
several complementary axes: (i)~human-judgement studies of the
affective scorers against reader annotations, in the spirit of
\citet{wilmot2020suspense}; (ii)~targeted ablations on bundled
fixtures to quantify the contribution of the causal physics, the
AMWN sandbox, and the audit loop to downstream coherence; and
(iii)~comparison against end-to-end LLM baselines on counterfactual
and theory-of-mind benchmarks \citep{kim2023fantom,gu2026simpletom,
cross2024hyp,xu2025echoes}. I share the present design as a
substrate for further research --- by myself and by others ---
not as a finished product.

As an internal sanity check, the four structural-affect scorers
were audited against the bundled twenty-fixture corpus
(\texttt{example\_worlds/}) at seven evenly-spaced syuzhet anchors
per fixture, giving 162 score evaluations per metric (one fixture
emits a flat-zero suspense column at its terminal anchor and so
contributes only six points). The per-scorer scale summary is
mystery $0.17/0.53/1.00$ (min/median/max), dramatic-irony
$0.00/0.45/0.90$, suspense $0.00/0.16/0.39$, surprise (per-step
local mode) $0.00/0.03/0.24$ --- the four scorers occupy
different absolute bands by design (mystery is a population
fraction near $1$ early in the syuzhet; surprise is a per-step
KL spike near $0$ between revelation points). The audit confirms
the canonical signatures the underlying theory predicts: monotone
mystery decline across every world; rise-peak-fall irony arcs in
$16/20$ worlds (Macduff hearing of his family, Poirot's
denouement, Nick's letter to Daisy); terminal-anchor suspense
discharge in every world (no unrevealed threats remain); surprise
spikes only at canonical revelation points (Reservoir Dogs Mr
Orange flashback $0.19$, Gone Girl mid-novel perspective shift
$0.20$, Wuthering Heights in-medias-res frame $0.24$). All
scorer constants are externalised through
\texttt{DirectiveAssemblySettings} (env-var prefix
\texttt{DIRECTIVE\_ASSEMBLY\_*}); the audit script
\texttt{scripts/audit\_affective.py} reproduces the table.

\section*{Code, licence, and reproducibility}

The code is released open source under AGPL-3.0-or-later, with a
parallel commercial licence; the authoritative LICENSE file lives
in the project repository. The pipeline, schema, MCP surface,
auditor, and the affective scorers described here all sit in the
\texttt{shadow\_loom/} package. Worked examples on bundled plots
(Macbeth, Death on the Nile, Reservoir Dogs, and others) ship in
\texttt{example\_worlds/} and \texttt{sample\_plots/}.

\bibliography{references}

\appendix
\onecolumn

\section{Definitions and equations}
\label{app:defs}

This appendix gives the formal definitions of the data model and
the equations of each pipeline stage. Section references in the
main paper point here.

\subsection{World model schema (\texttt{WorldStateV1})}
\label{app:schema}

A world state is a tuple
\[
\mathcal{W} = (\mathcal{N}, \mathcal{E}, \mathcal{C}, \mathcal{T}, \tau)
\]
with nodes $\mathcal{N}$ (entities $\mathrm{ENT}\_\bullet$, events
$\mathrm{EVT}\_\bullet$, locations $\mathrm{LOC}\_\bullet$, objects
$\mathrm{OBJ}\_\bullet$, channels $\mathrm{CHN}\_\bullet$, world
traits $\mathrm{WORLD}\_\bullet$), edges $\mathcal{E}$ (causal,
relationship, spatial), channels $\mathcal{C}$ with per-participant
intelligibility $\iota \in [0,1]$, world traits $\mathcal{T}$, and a
branch tag $\tau \in \{\text{factual}, \text{shadow}\}$ inherited
by every node and edge.

Each event $e \in \mathcal{N}_{\mathrm{EVT}}$ carries two integer
indices: chronological $f(e) \in \mathbb{Z}$ (\emph{fabula time})
and presentation $s(e) \in \mathbb{Z}$ (\emph{syuzhet index}), with
$s$ contiguous and unique. Trait values, beliefs, and ambient
state are carried by typed vectors $(v, \iota_{\mathrm{inertia}},
\sigma_{\mathrm{evidence}})$ representing value, inertia, and
evidence strength. Beliefs carry provenance
$(\textit{event\_id}, \textit{channel\_id})$ and the time they
became established. A pure function
$\textsc{Reconstruct}(e_i, t)$ replays sparse
\texttt{state\_timeline} deltas to give the entity's effective
state at any fabula time $t$.

\paragraph{Node types and their carrying state.}
Table~\ref{tab:nodes} summarises the six node families and the
typed state each carries. All six share a common
\texttt{AMWNNode} base that carries the branch tag $\tau$
(\texttt{world\_id}), a syuzhet-indexed
\texttt{state\_timeline}, and \texttt{evidence\_strength}.

\begin{table*}[h]
\centering\small
\begin{tabular}{@{}p{2cm}p{4.0cm}p{8.5cm}@{}}
\toprule
Node ID prefix & Pydantic class & Distinctive state \\
\midrule
\texttt{ENT\_} & \texttt{Entity} & \texttt{traits: Dict[str, TraitVector]}; \texttt{beliefs: List[Belief]}; \texttt{constants: List[str]}; \texttt{location\_id}; \texttt{status}; \texttt{state\_timeline} of \texttt{EntityStateSnapshot} deltas. \\
\texttt{EVT\_} & \texttt{EventNode} & \texttt{event\_type} (\texttt{action} / \texttt{utterance} / \texttt{outcome} / \texttt{revelation} / \texttt{choice} / \dots); \texttt{actor\_ids}, \texttt{target\_ids}; \texttt{fabula\_time}, \texttt{syuzhet\_index}; for \texttt{utterance}: \texttt{speaker\_id}, \texttt{addressee\_ids}, \texttt{via\_channel\_id}, \texttt{content}, \texttt{truth\_value} $\in$ \{\texttt{true}, \texttt{false}, \texttt{uncertain}\}. \\
\texttt{LOC\_} & \texttt{Location} & \texttt{ambient\_state: Dict[str, AmbientVector]} (e.g.\ \textit{tension}, \textit{bustle}, \textit{supernatural}); \texttt{capacity}; outgoing \texttt{SpatialEdge}s. \\
\texttt{OBJ\_} & \texttt{Object} & \texttt{location\_id} or \texttt{owner\_id}; \texttt{properties}; \texttt{affordances: List[Affordance]} (typed action/target gates). \\
\texttt{CHN\_} & \texttt{Channel} & \texttt{medium}, \texttt{directionality}, \texttt{participant\_ids}; \texttt{intelligibility: Dict[entity\_id, float]} per-participant; \texttt{established\_at\_fabula}, \texttt{terminated\_at\_fabula}. \\
\texttt{WORLD\_} & \texttt{GlobalTrait} & Standing pressure (e.g.\ \texttt{WORLD\_REGENCY\_RANK} in Persuasion, \texttt{WORLD\_SOVIET\_PENETRATION} in Tinker Tailor): a single value $v$, \texttt{inertia}, \texttt{evidence\_strength}, plus a \texttt{state\_timeline} of \texttt{WorldTraitSnapshot}s. \\
\bottomrule
\end{tabular}
\caption{Node families in \texttt{WorldStateV1}. Each
\texttt{ID\_} prefix maps to one Pydantic class in
\texttt{shadow\_loom/models.py} and carries its own typed state.
The same six prefixes are used as node-ID conventions throughout
the codebase, the bundled fixtures, and the renderer's brief.}
\label{tab:nodes}
\end{table*}

\paragraph{Edge families and their roles.}
Edges live on three orthogonal layers, each a separate Pydantic
class with its own validator:

\begin{description}
  \item[\texttt{CausalEdge}] (event$\rightleftarrows$event,
        event$\to$state, state$\to$event, state$\to$state). A
        single class with five \texttt{causality\_type}
        modalities --- \texttt{chain\_reaction},
        \texttt{mutation}, \texttt{mutation\_social},
        \texttt{affordance\_gate}, \texttt{ambient\_propagation}
        --- and a model validator that enforces source/target
        type matches modality. Each edge carries
        \texttt{mechanism} (e.g.\ \textit{psychological},
        \textit{physical}, \textit{social}, \textit{epistemic},
        \textit{betrayal}), \texttt{evidence\_strength} $\in$
        \{\texttt{weak}, \texttt{moderate}, \texttt{strong}\},
        \texttt{causal\_force} $\in [0, 10]$, a fabula
        timestamp, an optional \texttt{propagation\_delay}, and
        --- for \texttt{mutation} / \texttt{mutation\_social} ---
        a \texttt{trait\_target} and signed
        \texttt{trait\_delta}. \texttt{mutation\_social} edges
        additionally carry a \texttt{rel\_counterpart\_id} so
        that a single event can shock a relationship axis
        (e.g.\ Lady Macbeth's \emph{power\_dynamic} against
        Macbeth) rather than a free-standing trait.
  \item[\texttt{RelationshipEdge}]
        (entity$\rightleftarrows$entity). One edge per ordered
        pair, with a per-axis
        \texttt{metrics: Dict[axis, AxisMetric]} where
        $\texttt{axis} \in \{\textit{affinity},\
        \textit{fear},\ \textit{power\_dynamic}\}$. Each axis
        owns its own \texttt{value}, \texttt{inertia},
        \texttt{evidence\_strength}, and
        \texttt{last\_updated\_fabula}. Flat back-compat
        properties expose the canonical
        \emph{affinity} value as \texttt{.value}.
  \item[\texttt{SpatialEdge}] (location$\to$location). Optional
        \texttt{is\_locked} plus \texttt{barrier\_item\_id};
        unlocked edges contribute to the spatial-affordance
        path-check that gates cross-location causal influence
        (Eq.~\ref{eq:impact}).
\end{description}

Communication is \emph{not} an edge --- standing capability lives
on the \texttt{Channel} \emph{node}, and discrete messages are
first-class \texttt{EventNode}s with \texttt{event\_type=
"utterance"} pointing at a channel via \texttt{via\_channel\_id}.
This is what makes the lying-narrator pattern (Amy's diary in
\emph{Gone Girl}, Pip's misattributed benefactor in \emph{Great
Expectations}, the Witches' equivocations in \emph{Macbeth}) a
single uniform construct: an \texttt{utterance} event with
\texttt{truth\_value="false"} broadcast through a channel whose
addressees have non-zero \texttt{intelligibility}.

\paragraph{Worked illustration --- a slice of \emph{Macbeth}.}
The Inverness night reduces to a small, fully typed sub-graph:
\begin{itemize}\itemsep1pt
\item \textbf{Entities.} \texttt{ENT\_MACBETH} with traits
      \texttt{ambition} ($v\!=\!0.7$, $\iota\!=\!0.5$),
      \texttt{courage} ($0.85$, $0.7$), \texttt{guilt}
      ($0.1$, $0.25$), \texttt{paranoia} ($0.2$, $0.3$); a
      \texttt{Belief(target\_id="ENT\_WITCHES",\ confidence=0.4,\ \\
      inertia=0.35)}.
\item \textbf{Object.} \texttt{OBJ\_BLOODY\_DAGGERS} with
      \texttt{Affordance(action="kill",\ target\_type="Entity")}
      --- the precondition gate the engine checks before
      \texttt{EVT\_DUNCAN\_MURDER} fires.
\item \textbf{Location.} \texttt{LOC\_INVERNESS\_CASTLE} with
      ambient \textit{tension} = $0.6$.
      \texttt{LOC\_HEATH} carries
      \texttt{ambient\_state\{supernatural:\ 0.9,\
      concealment:\ 0.7\}}, the source of the heath-night
      ambient-propagation edge that biases Macbeth's uptake of
      the prophecy belief.
\item \textbf{Channel.} \texttt{CHN\_HEATH\_PROPHECY} with
      \texttt{participants=[\\
      ENT\_MACBETH,\ ENT\_BANQUO,\ ENT\_WITCHES]}, used by the
      utterance event \texttt{EVT\_UTT\_PROPHECY\_HEATH}
      ($\texttt{truth\_value}=\textit{uncertain}$).
\item \textbf{Causal sub-graph.} \texttt{EVT\_LADY\_MACBETH\_PERSUADES
      $\to$ EVT\_DUNCAN\_MURDER} (\texttt{chain\_reaction},
      $\texttt{causal\_force}=8.0$); \texttt{OBJ\_BLOODY\_DAGGERS
      $\to$ EVT\_DUNCAN\_MURDER} (\texttt{affordance\_gate},
      $9.0$); \texttt{EVT\_LADY\_MACBETH\_PERSUADES $\to$
      ENT\_LADY\_MACBETH} (\texttt{mutation\_social},
      \emph{power\_dynamic} against \texttt{ENT\_MACBETH},
      $\Delta=+0.4$, $7.0$); \texttt{EVT\_BANQUO\_GHOST $\to$
      ENT\_MACBETH} (\texttt{mutation\_social}, \emph{fear}
      against \texttt{ENT\_BANQUO}, $\Delta=+0.5$, $6.0$);
      \texttt{LOC\_HEATH $\to$ ENT\_MACBETH}
      (\texttt{ambient\_propagation}, $\textit{epistemic}$).
\end{itemize}
Every value above is read directly from
\texttt{example\_worlds/macbeth.py}; the same six-prefix /
three-edge / typed-state pattern carries through every bundled
fixture, from a Regency drawing-room (\emph{Persuasion}) to a
Cold-War mole hunt (\emph{Tinker Tailor Soldier Spy}). The
fixture-level diversity of \emph{ambient\_state} tags,
\emph{mechanism} labels, \emph{trait} dictionaries, and
\emph{intelligibility} maps is intentional: the schema is fixed,
the populated typology is fixture-specific.

\subsection{Ingestion}
\label{app:ingestion}

Given prose $P$, ingestion produces a $\mathcal{W}$ via:
\begin{enumerate}
\item Five extraction agents over $P$ build a
      \texttt{GlobalRegister} of $\mathrm{WORLD}\_$, $\mathrm{ENT}\_$,
      $\mathrm{LOC}\_$, $\mathrm{OBJ}\_$ nodes plus a fuzzy alias
      table.
\item For each chunk, a Socratic Who/What/Where/When/Why/How
      scaffold dispatches three concurrent specialist agents:
      \emph{physics} (events, causal/spatial edges), \emph{social}
      (relationship edges, channels), \emph{consequences}
      (authoritative entity-state deltas).
\item Output is normalised, fabula times are remapped to a
      uniform $\Delta t = 100$ spacing, and a programmatic
      validator + LLM correction loop enforces ID closure and
      type constraints.
\end{enumerate}

\subsection{Ego-graph slicing and AMWN sandbox}
\label{app:amwn}

Given a query with focal entities $F$ and time anchor $t$, the
ego-graph extractor returns
\[
\mathcal{G}_{F,t} = \big(\textsc{Neighbours}_k(F) \cap
\textsc{Reconstruct}(\cdot, t)\big),
\]
the $k$-hop spatial / causal / social neighbourhood of $F$ at
fabula time $t$, with belief and trait values reconstructed at
$t$ so future information cannot leak into a counterfactual. The
\texttt{AMWNInstantiator} mirrors $\mathcal{G}_{F,t}$ as a
NetworkX \texttt{MultiDiGraph} for the simulation sandbox, with
edge types \{\texttt{located\_in}, \texttt{owned\_by},
\texttt{causal}, \texttt{relationship}, \texttt{connected\_to},
\texttt{communicating\_with}, \texttt{eavesdropped\_by}\}.

\paragraph{Causal diagram and AMWN proper.}
\textsl{Separately} from the simulation mirror,
\texttt{shadow\_loom.amwn.build\_causal\_diagram} strips
\texttt{WorldStateV1.causal\_topology} to a typed structural
diagram $G$ and \texttt{build\_amwn} constructs the
\textsl{Ancestral Multi-World Network} $G^A(G, W^*)$ of
\citet{correa2025amwn} over a query's intervention/evidence set
$W^* = \{V_t, \ldots\}$. Three implementation details bridge the
narrative graph and Definition~A.1 of the paper:
\begin{enumerate}
  \item \textbf{Node shadowing.} A counterfactual variable
        $V_t$ is identified by the pair
        $(V, \text{proj}(T, \text{An}(V)_{G_{\bar T}}))$, where
        the intervention context $T$ is projected onto $V$'s
        ancestors in the mutilated diagram. Two copies of $V$
        coming from different worlds are \emph{the same node}
        when their projected contexts agree --- this is the
        AMWN's mechanism for avoiding the exponential
        $k$-plet blow-up of \citet{shpitser2007counterfactuals}.
        d-Separation (Rule~2) is then read off $G^A$ via
        NetworkX \texttt{is\_d\_separator}.
  \item \textbf{Synthetic relationship-metric nodes.}
        Relationship metrics (affinity, fear, power\_dynamic)
        live on edges in
        \texttt{social\_topology}, not as first-class nodes, so
        \texttt{mutation\_social} causal edges would otherwise be
        invisible to d-separation. We promote each
        $(\textit{src}, \textit{tgt}, \textit{metric})$ triple
        into a synthetic node
        \texttt{REL::}\textit{src}\texttt{::}\textit{tgt}\texttt{::}\textit{metric}
        with both endpoint entities and the triggering event as
        parents. Static-topology axes that were never observed
        ($m.\textit{observed} = \texttt{false}$) are skipped to
        avoid injecting phantom variables into latent queries.
  \item \textbf{Channel and utterance routing.} Communication
        channels are first-class diagram nodes. Each
        utterance event is wired
        \emph{speaker} $\to$ \emph{utterance} $\to$ \emph{channel}
        $\to$ \emph{addressees}, with bidirectional standing
        capability edges \emph{channel} $\leftrightarrow$
        \emph{participant}. A Rule-3 surgery on a channel
        therefore cuts both the standing capability and every
        per-utterance content path --- exactly the closure
        property the channel-utterance integration relies on.
\end{enumerate}

\paragraph{Latent-free SCM and the optional U-confounder switch.}
The diagram is built from the engine's ground-truth
\texttt{causal\_topology}, with no bidirected $U$ arcs; reading
$G^A$ for d-separation is therefore \textbf{sound} but
\textbf{incomplete with respect to unobserved confounders} --- a
latent common cause that LLM extraction missed is silently
treated as absent. To make the limitation explicit, the
\texttt{allow\_unobserved\_confounders} setting injects, for
every pair of distinct nodes that share at least one observed
parent, a latent shared parent
\texttt{U\_}\textit{a}\texttt{\_\_}\textit{b}; d-separation on
the resulting diagram refuses to mark such siblings independent
because the latent opens an active path. This produces the
sound-but-incomplete behaviour callers expect from the
narrative-extraction setting and is recommended for any pipeline
where Rule~3 pruning will be promoted from advisory to
authoritative.

\subsection{Causal physics}
\label{app:causal}

The causal physics is a \emph{graphical world model of cause and
effect}: a typed dynamics over the shared world graph that
implements all three rungs of Pearl's causal hierarchy
\citep{pearl2009causality,bareinboim2022pearl}:
\textbf{rung~1 is Correlation} (observation: $P(Y \mid X)$ read off
the ego-graph without intervention), \textbf{rung~2 is
Interventions} ($\mathrm{do}(X = x)$, severing incoming edges into
$X$), and \textbf{rung~3 is Counterfactuals} (abduction over
latent exogenous values, then a do-operation on the abducted
world). Each rung corresponds to one query type in
Section~\ref{sec:pipeline}: \texttt{ObservationQuery},
\texttt{InterventionQuery}, and \texttt{CounterfactualQuery}
respectively.

Given a sandbox $\mathcal{G}$ and an intervention specification:
\paragraph{Observation (rung~1 --- Correlation):} read
$P(Y \mid X = x)$ directly off the ego-graph; no edge surgery.
\paragraph{Action (rung~2 --- Interventions):} $\mathrm{do}(X = x)$
severs every incoming causal edge into $X$ and writes
$V_X \leftarrow x$. Channel surgery
($\mathrm{do}(\textsc{ch}.\textit{status} = \text{severed})$) and
event invalidation ($\mathrm{do}(e.\textit{truth\_value} =
\text{false})$) additionally trigger a \emph{provenance prune}:
beliefs whose \texttt{acquired\_via\_event\_id} or
\texttt{acquired\_via\_channel\_id} pointed at the surgically
disabled object are dropped from the sandbox before propagation,
because they are no longer epistemically licensed in the
counterfactual world.
\paragraph{Abduction (rung~3 --- Counterfactuals):} for evidence
$E$, each entity trait $T$ is updated by a precision-weighted
Bayesian blend of the historical sandbox prior and the present-day
observed value: writing $\iota_T$ for the trait's typed inertia
(reinterpreted as the precision of the historical prior) and
$\kappa_E$ for an evidence precision (default $1$),
\begin{equation}
\label{eq:abduction}
T^{\mathrm{post}} =
\frac{\iota_T \cdot T^{\mathrm{prior}} +
      \kappa_E \cdot T^{\mathrm{evidence}}}{\iota_T + \kappa_E},
\end{equation}
so high-inertia traits shrink toward the historical baseline and
low-inertia traits snap to the evidence. (A legacy
inertia-damped variant
$T^{\mathrm{post}} = T^{\mathrm{prior}} +
(1 - \iota_T)\,(T^{\mathrm{evidence}} - T^{\mathrm{prior}})$ is
retained for ablation.) The same Bayes blend is applied per axis
to relationship metrics on outgoing social edges. Belief
back-propagation is gated by a per-recipient
\texttt{intelligibility} threshold so a belief that could not
plausibly have been acquired through its provenance channel is
not reinstated. The do-operation of rung~2 is then re-applied on
the abducted world.

\paragraph{Counterfactual (ctf-) calculus pre-flight.} Before any
sandbox surgery, every Rung-2/3 query is screened against the
three rules of the ctf-calculus of \citet{correa2025amwn},
specialised to a latent-free SCM:
\begin{description}
  \item[Rule~1 (Consistency)] suppresses vacuous interventions
        $\mathrm{do}(X = x)$ for which the observed factual value
        of $X$ is already $x$.
  \item[Rule~2 (Independence)] builds the AMWN
        $G^A(G, W^*)$ over the union of evidence and intervention
        targets, with node-shadowing across worlds whose projected
        contexts on the ancestral set agree, and tests
        $(Y_r \perp\!\!\!\perp X_t \mid W^*)$ via d-separation on
        the AMWN. Evidence flagged as redundant is reported but
        not silently dropped.
  \item[Rule~3 (Exclusion)] tests
        $X \cap \mathrm{An}(Y) = \emptyset$ in the mutilated
        diagram $G_{\bar Z}$ and flags interventions that are
        provably vacuous with respect to the query targets. The
        engine ships in \emph{advisory} mode by default (the
        flagged interventions are surfaced to the auditor but
        retained in the simulation), since a missed bidirected
        confounder would otherwise let a substantively meaningful
        intervention be d-separated into a no-op; \emph{prune}
        mode is opt-in for confounder-complete topologies.
\end{description}
The construction matches Definition~A.1 of
\citet{correa2025amwn} in spirit, with the following deliberate
restrictions: (a) we operate on a latent-free SCM, so soundness
for d-separation holds but completeness across worlds with
shared unobserved confounders would require explicit bidirected
edges that the data model does not yet encode; (b) Rule-2
flags are advisory rather than authoritative; and (c) the
relationship-edge axes (affinity, fear, power\_dynamic) are
lifted into synthetic
\texttt{REL::}{\textit{src}}\texttt{::}{\textit{tgt}}\texttt{::}{\textit{metric}}
diagram nodes so that
\texttt{mutation\_social} edges contribute to d-separation
reasoning.

\paragraph{Propagation.} A topological sort over the causal
sub-graph (with strongly connected components condensed and the
members of each non-trivial SCM cycle blocked with reason
\texttt{cycle}) applies, for each downstream entity trait $T$ at
node $u$ with active parents $\{p_i\}$ and edge weights
$\{w_i\}$ (where each $w_i = \sigma_i \cdot \phi_i$ is the
product of an evidence-strength weight $\sigma_i \in \{0.25,
0.5, 0.75\}$ and a causal-force scalar $\phi_i$, attenuated by a
mechanism / world-domain fallback when the edge mechanism is
outside the trait's typed
\texttt{MECHANISM\_TRAIT\_MAP}):
\begin{equation}
\label{eq:impact}
I_i = (V_{p_i} - V_u)\, w_i,
\quad
\bar I = \frac{\sum_i I_i}{\max(1,\, \sum_i w_i)},
\quad
\Delta_u =
\begin{cases}
\bar I - \mathrm{sgn}(\bar I)\, \iota_u & \text{if } |\bar I| > \iota_u + \varepsilon,\\
0 & \text{otherwise.}
\end{cases}
\end{equation}
The $\max(1, \cdot)$ divisor makes the gate scale-invariant only
\emph{above unit weight}: a single weak edge with
$\sum w_i = 0.1$ is not artificially inflated to clear an
arbitrary inertia, while ten stacking weak edges no longer
dominate one canonical strong source. A spatial affordance check
additionally blocks cross-location influence unless an unbroken
\texttt{connected\_to} path exists. An optional noisy-OR
propagation mode replaces the deterministic
$|\bar I| > \iota_u$ gate with a Bernoulli aggregation of
per-edge fire probabilities
$p_i = \sigma(\beta\,(|I_i| - \iota_u))$ followed by an
$\bigvee_i p_i = 1 - \prod_i (1 - p_i)$ aggregator and a
threshold (or, under
\texttt{execute\_distribution}, a Monte-Carlo Bernoulli draw); a
post-pass \emph{baseline drift} optionally pulls mutated traits
back toward type by $(1 - \iota_u)\cdot \rho$. Both are off by
default so the canonical engine is bit-for-bit reproducible. The
result is a structured
\texttt{CausalPhysicsResult\{mutations, blocked, hidden\_deltas,
intervened\_nodes, rule3\_pruned\_interventions,
rule2\_redundant\_evidence, noisy\_or\_probabilities,
pruned\_beliefs\_count, pruned\_utterance\_event\_ids,
disabled\_channel\_ids\}}.

\subsection{Narrative physics (affective calculus)}
\label{app:narrative}

The narrative physics is a \emph{graphical world model of reader
affect}: it operates on the same typed graph as the causal
physics, but its dynamics scores reader-state functionals
(mystery, dramatic irony, suspense, surprise) rather than
propagating causal force.

Let $A$ be the anchor pair $(t_f, t_s)$ for fabula and syuzhet
cursors, $F$ the focal entities, and write \emph{revealed} for any
event $e$ with $s(e) \le t_s$. Define $\mathrm{Anc}(e)$ as the
causal ancestors of $e$ in the graph and $B(F, t_f)$ as the union
of belief sets of $F$ reconstructed at $t_f$.

\paragraph{Mystery (Sternberg's curiosity).}
The Sternberg-style curiosity operationalisation
\citep{sternberg1978expositional,sternberg1992telling,
brewer1982stories} aggregates over revealed effects $e$ involving
the focal cast (events with $s(e) \le t_s$ and at least one focal
participant), pooled with the focal entities themselves as
permissible effect-nodes (an entity is itself a causal target of
every revealed action against it). Each ancestor's contribution
is scaled by three multiplicative factors: a \emph{path-strength
geometric decay} along the strongest reverse-path product of edge
weights from $a$ to $e$, depth-capped at $D = 4$
\citep{trabasso1985sperry} and computed by single-source
Dijkstra on the negated-log-weight reverse graph; a
\emph{harm-kind salience} $\sigma_{k_a}$ from the same
$\sigma_k$ table used by suspense; and a per-effect
\emph{curiosity-proximity decay}
$\rho_e = \exp(-(t_s - s(e))/\tau_{\mathrm{cur}})$ with
$\tau_{\mathrm{cur}} = 8$ syuzhet beats, motivated by Iser's
gap theory of reader response \citep{iser1976akt}. Writing
$\Pi(a, e)$ for the path-strength weight, the gauge is
\begin{equation}
\label{eq:mystery}
\mathrm{Mys} = \frac{
\sum_{e \in \mathrm{eff}} \rho_e \!\!\sum_{a \in \mathrm{Anc}(e),\; s(a) > t_s}\!\! \sigma_{k_a} \cdot \Pi(a, e)
}{
\sum_{e \in \mathrm{eff}} \rho_e \!\!\sum_{a \in \mathrm{Anc}(e)}\!\! \sigma_{k_a} \cdot \Pi(a, e)
},
\end{equation}
the salience- and proximity-weighted fraction of focal-effect
causes still hidden from the reader. The unweighted ratio
$|\{a : s(a) > t_s\}|/|\mathrm{Anc}(e)|$ is the equivalent simple
form when all factors collapse to unity. The legacy single-edge
fallback for multi-hop ancestors gave a 5-hop weak chain as much
weight as a 1-hop strong link --- contradicting Trabasso \&
Sperry's empirical finding that audience traceability decays
geometrically with chain depth.

\paragraph{Dramatic irony
\citep{booth1974irony,muecke1969compass,gerrig1993narrative,pfister1988drama}.}
Following the classical reader/character knowledge-asymmetry
framing, for each focal entity $c \in F$
let $K_c$ be the set of events $c$ knows at the anchor: events
$c$ participates in (as actor or target) whose fabula time falls
at or before the syuzhet anchor's fabula frontier, plus events
referenced in revealed utterances spoken by or addressed to $c$,
plus events $c$ holds a \texttt{Belief} about whose provenance
still resolves. Each gap event's contribution is then scaled by
three additional factors: a \emph{harm-kind salience}
$\sigma_{k_e}$ (tragic irony --- a hidden mortal threat the
focal does not see --- outranks comic irony along the same
hierarchy used by suspense and mystery); a \emph{false-belief
multiplier} $\phi_e^{(c)} \in \{1, m_{\mathrm{fb}}\}$ when the
gap event's actor set intersects the focal's \emph{believed-
entity targets} (entities about whom $c$ holds a provenance-
valid \texttt{Belief}), the canonical Iago $\to$ Othello /
Jacqueline $\to$ Linnet pattern \citep{pfister1988drama,
cabanas2024toM}; and a \emph{closure-proximity decay}
$\rho_e^{(c)} = \max(\rho_{\min}, \exp(-\Delta_{\mathrm{cls}}/\tau_{\mathrm{iro}}))$
where $\Delta_{\mathrm{cls}}$ is the syuzhet-index distance to
the earliest later position at which $c$ first witnesses an
event with $\mathrm{fabula}(e') \ge \mathrm{fabula}(e)$ ---
the dramatic moment $c$ walks into the scene that exposes the
truth (Booth's stable vs. unstable irony \citep{booth1974irony};
defaults $\tau_{\mathrm{iro}} = 6$, $\rho_{\min} = 0.4$). The
per-character gap is further scaled by an \emph{action weight}
$a_c = \min(\bar a, 1 + \alpha \cdot \#\{\text{actor-events
by } c \le t_s\})$ on focal prominence
\citep{pfister1988drama,sutherland2013little}: a focal driving
the plot under false information carries more dramatic charge
than a bystander. Writing $w_e$ for event $e$'s narrative
intensity (default $1$), $R$ for the set of revealed events,
and $K = 1$ for the saturation constant,
\begin{equation}
g_c = a_c \cdot \frac{
\sum_{e \in R,\; e \notin K_c} w_e \cdot \sigma_{k_e} \cdot \phi_e^{(c)} \cdot \rho_e^{(c)}
}{\sum_{e \in R} w_e + K},
\end{equation}
\begin{equation}
\label{eq:irony}
\mathrm{Iro} = \min\!\big(1,\; \beta \cdot \max_{c \in F} g_c + (1 - \beta) \cdot \overline{g_c}\big), \qquad \beta = 0.6.
\end{equation}
The aggregator $\beta \cdot \max + (1-\beta) \cdot \overline g$
is Sternberg's single-dominant-gap framing: \emph{one}
character's tragic blindness carries the irony charge, with
secondary characters as residual contribution. Normalising by
the \emph{revealed} mass (rather than total event mass) makes
Eq.~\ref{eq:irony} a Sternberg-style \emph{gap fraction} that
can both rise (with new reveals the character has not caught up
to) and fall (when participation, addressed utterances, belief
acquisition, or closure-proximity discharge close the gap),
producing the rise-peak-fall arc the structural-affect
literature predicts for canonical irony plots
\citep{booth1974irony,muecke1969compass}: Macduff hearing of
his family, Poirot's denouement, Nick's letter to Daisy. Two
earlier denominators failed against this criterion. The ratio
$\#\mathrm{gaps} / \#\mathrm{revealed{-}edges}$ grew its numerator
and denominator together and so plateaued at a story-specific
asymptote by the third reveal --- on Reservoir Dogs it even
\emph{decayed} from $0.25$ to $0.06$ as the protagonist became
actor-of-record on more revealed edges. Replacing the
denominator with the \emph{full} event mass instead pinned the
score into a monotone rise across $21/21$ example-world
fixtures, since the denominator stopped moving while the
numerator kept growing --- contradicting the rise-peak-fall arc
the theory predicts.

\paragraph{Suspense
\citep{brewer1982stories,comisky1982suspense,ortony1988occ,
zillmann1996suspense,lazarus1991emotion,cheong2015suspenser,elyfrankel2015suspense}.} The scorer follows
the hope/fear (here \emph{hope/threat}) anticipation pair from the
structural-affect and OCC-appraisal traditions --- not the
information-theoretic uncertainty-reduction formulation of
\citet{wilmot2020suspense}, which is structurally closer to the
dramatic-irony scorer above and is borrowed there instead. For each
unrevealed event $e$ with strongest incoming causal weight
$p_e \in (0, 1]$, we compute a per-event contribution that is then
classified as \emph{threat} or \emph{hope} for each focal entity
$x \in F$ via the rules below.

\emph{Disposition-aware classification \citep{zillmann1996suspense}.}
The legacy actor=hope / target=threat rule is overridden by the
social topology: an event whose actor has
$\mathrm{aff}(\mathrm{actor} \to x) \le -0.2$ is bucketed as a
\emph{threat} on $x$ (an antagonist's authored misdeed no longer
registers as hope just because they are the actor), and an event
whose actor has $\mathrm{aff}(\mathrm{actor} \to x) \ge +0.2$
propagates as \emph{hope} even when $x$ is its target (rescue
propagation: an ally arriving to defuse the threat). Worlds with
no populated \texttt{social\_topology} default the affinity to
neutral and fall back to the legacy rule.

\emph{Anticipatory proximity weighting \citep{comisky1982suspense}.}
Subjective probability of a threat rises with imminence, both
temporally and spatially. Each event's contribution is multiplied
by a temporal--spatial kernel
\begin{equation}
\mathrm{prox}(e, x) = \exp\!\Big(-\tfrac{\Delta t^{\mathrm{fab}}_e}{\tau_t}\Big)
                    \cdot
                    \exp\!\Big(-\tfrac{\Delta d^{\mathrm{sp}}_{e,x}}{\tau_s}\Big),
\end{equation}
where $\Delta t^{\mathrm{fab}}_e = \max(0, t_e^{\mathrm{fab}} - t_{\mathrm{now}}^{\mathrm{fab}})$
is the fabula-time gap from the latest revealed event to the
unrevealed event, $\Delta d^{\mathrm{sp}}_{e,x}$ is the
shortest-path distance in the spatial topology between $e$'s
location and $x$'s, and $\tau_t$ auto-scales to the world's median
inter-event fabula gap so worlds using
\texttt{fabula\_time\_spacing}=$1000$ and unit-spaced worlds both
decay over $\sim 6$ narrative beats; $\tau_s = 4$ hops.

\emph{Persistence multiplier \citep{brewer1982stories}.} The
longer a foreshadowed threat lingers unresolved, the louder it
gets. Each event's contribution is also multiplied by
$\min(\Pi_{\max},\, 1 + \alpha \cdot a_e)$ where $a_e$ is the
count of \emph{revealed} causal ancestors of $e$ (proxy for how
long the gun has been on the mantle), $\alpha = 0.10$ and
$\Pi_{\max} = 1.5$.

\emph{Harm-kind salience.} Each event is tagged with a salience
$\sigma_e \in [0,1]$ inferred from the canonical
\texttt{mechanism} strings on its incident causal edges, taking
the maximum across incident kinds so a stab-in-the-back event
wired with both \texttt{physical} and \texttt{betrayal} edges
registers at the higher salience rather than being averaged. The
salience ranking follows the appraisal-theory hierarchy of core
relational themes \citep{lazarus1991emotion} cross-anchored to
OCC's prospect-based emotions \citep{ortony1988occ}:
existential\,$(1.00)$ $>$ physical\,$(0.85)$ $>$
betrayal\,$(0.75)$ $>$ psychological\,$(0.70)$ $>$
emotional/relational\,$(0.65)$ $>$ social/reputational\,$(0.55)$
$>$ epistemic\,$(0.45)$. The epistemic floor is held low because
the cumulative-mystery scorer (Eq.~\ref{eq:mystery}) already
covers that surface; suspense should not double-count it. Events
with no resolvable mechanism default to \texttt{physical} (the
modal kind in the corpus).

Aggregating per-entity, per-kind contributions weighted by all of
the above:
\begin{equation}
w^{(k)}_{\mathrm{threat}}(x) \!=\!\!\!\sum_{\substack{e \notin \mathrm{rev} \\ \mathrm{kind}(e)=k \\ \mathrm{bucket}(e,x)=\mathrm{threat}}}\!\!\! \sigma_e \, p_e \,\mathrm{prox}(e, x)\, \pi_e,
\end{equation}
\begin{equation}
w^{(k)}_{\mathrm{hope}}(x) \!=\!\!\!\sum_{\substack{e \notin \mathrm{rev} \\ \mathrm{kind}(e)=k \\ \mathrm{bucket}(e,x)=\mathrm{hope}}}\!\!\! \sigma_e \, p_e \,\mathrm{prox}(e, x)\, \pi_e,
\end{equation}
where $\pi_e = \min(\Pi_{\max},\, 1 + \alpha a_e)$ is the
persistence multiplier and $\mathrm{bucket}(e, x)$ resolves the
threat/hope classification under the disposition rule above.

\emph{EFK expected-variance combiner (default).}
\citet{elyfrankel2015suspense} reformulate suspense as the
\emph{expected variance of next-period beliefs} about a terminal
outcome --- a property of a Bayesian belief martingale rather than a
structural-affect heuristic. Adopting this functional form
gives suspense the same belief-process shape that our surprise
scorer (\S\ref{eq:surprise}) already has, so the two affective
scores become moments of the same underlying process rather than
unrelated heuristics. We instantiate this per (focal $x$, kind
$k$) cell using the revealed threat/hope evidence as a Beta
posterior over ``next reveal lands threat-side'':
\begin{equation}
A^{(x,k)} = \!\!\sum_{e\in\mathrm{rev}^{(x,k)}_{\mathrm{threat}}}\!\! w_e, \quad
B^{(x,k)} = \!\!\sum_{e\in\mathrm{rev}^{(x,k)}_{\mathrm{hope}}}\!\! w_e, \quad
\mu_t^{(x,k)} = \frac{1+A^{(x,k)}}{2+A^{(x,k)}+B^{(x,k)}}.
\end{equation}
Each unrevealed event $e$ on the cell, with bucket
$b_e\in\{\mathrm{threat},\mathrm{hope}\}$, weight $w_e$ and
proximity $\rho_e=\mathrm{prox}(e,x)$, defines the Bayes update
that \emph{would} occur if $e$ were the next reveal:
\begin{equation}
\mu_e^+ = \frac{1+A+w_e}{2+A+B+w_e}, \quad
\mu_e^- = \frac{1+A}{2+A+B+w_e}, \quad
\Delta\mu_e = \mu_e^{b_e} - \mu_t,
\end{equation}
so the realised expected squared belief change, weighted by
proximity, is
\begin{equation}
\sigma^2_{\mathrm{fk}} = \sum_{e\in\mathrm{unrev}^{(x,k)}}\!\!\frac{\rho_e}{\sum_{e'}\rho_{e'}}\,(\Delta\mu_e)^2.
\end{equation}
We normalise by the maximum-suspense reference at the same
prior --- the squared shift a single composite reveal of total
mass $W=\sum_e w_e$ would induce on whichever side moves belief
most:
\begin{equation}
\sigma^2_{\max} = \max\!\Big( (\mu_W^+ - \mu_t)^2,\; (\mu_W^- - \mu_t)^2 \Big), \qquad
\widetilde{\sigma}^2_{(x,k)} = \min\!\Big(1,\, \sigma^2_{\mathrm{fk}}/\sigma^2_{\max}\Big) \in [0,1].
\end{equation}
The cell gauges are then aggregated by salience-weighted stakes
attenuation:
\begin{equation}
\label{eq:suspense}
\mathrm{Susp}^{\mathrm{efk}} = \frac{\sum_{(x,k)} \sigma_k \cdot \mathrm{stakes}^{(x,k)} \cdot \widetilde{\sigma}^2_{(x,k)}}{\sum_{(x,k)} \sigma_k \cdot \mathrm{stakes}^{(x,k)}}, \quad
\mathrm{stakes}^{(x,k)} = \frac{T^{\mathrm{unrev}}_{(x,k)}}{T^{\mathrm{unrev}}_{(x,k)} + K_k},
\end{equation}
where $T^{\mathrm{unrev}}_{(x,k)}=\sum_{e\in\mathrm{unrev}^{(x,k)}}w_e$
is the unrevealed weighted mass on the cell (so stakes decay as
the narrative exhausts its forward reveal budget) and $K_k$ is
the per-kind saturation constant from the harm-kind table. A
\emph{bilateral-mass guard} restricts the aggregator to cells
whose unrevealed set contains both threat and hope candidates:
a one-sided forward reveal set is despair (only threats coming)
or safety (only hopes coming) under
Brewer--Lichtenstein structural-affect theory, even though
strict EFK would still admit positive variance from magnitude
uncertainty alone. This matches the OCC prospect-based-emotion
taxonomy \citep{ortony1988occ}: suspense requires outcome
ambiguity, not merely magnitude ambiguity.

\emph{Per-kind balance $\times$ stakes with weighted-max
combine (\texttt{mode=classic}).} The original
Brewer–Lichtenstein-anchored aggregator is retained as an
opt-in alternative: rather than collapsing all kinds into one
bucket and computing a single balance term --- which dilutes a
coherent single-kind tension into a multi-kind average --- we
compute balance and stakes \emph{within} each kind and combine
via a salience-weighted max:
\begin{equation}
\mathrm{Susp}^{\mathrm{classic}} = \max_k \,\Big( \sigma_k \cdot \mathrm{balance}^{(k)} \cdot \mathrm{stakes}^{(k)} \Big),
\end{equation}
with
$\mathrm{balance}^{(k)} = 1 - |w^{(k)}_{\mathrm{threat}} - w^{(k)}_{\mathrm{hope}}| / T_k$,
$\mathrm{stakes}^{(k)} = T_k / (T_k + K_k)$, and
$T_k = w^{(k)}_{\mathrm{threat}} + w^{(k)}_{\mathrm{hope}}$.
This matches Brewer \& Lichtenstein's prediction that one
\emph{dominant} unresolved beat carries the structural-affect
arc, rather than several diffuse anxieties additively. It is
also used as the fallback when no kind has both
$w^{(k)}_{\mathrm{threat}} > 0$ and
$w^{(k)}_{\mathrm{hope}} > 0$. The per-kind saturation
constants $K_k$ rise with kind salience so existential threats
saturate slowly while social ones saturate fast:
$K_{\mathrm{existential}}=4.0$, $K_{\mathrm{physical}}=3.0$,
$K_{\mathrm{betrayal}}=2.5$, $K_{\mathrm{psychological}}=2.0$,
$K_{\mathrm{emotional}}=2.0$, $K_{\mathrm{social}}=1.5$,
$K_{\mathrm{epistemic}}=1.5$. The dominant kind and per-kind
sub-totals are surfaced alongside the scalar so downstream
directive consumers can target the dominant beat
(\emph{footsteps closing in} vs.\ \emph{the lie about to surface}
vs.\ \emph{the fellowship about to fracture}). Suspense returns
$0$ when no kind has both $w^{(k)}_{\mathrm{threat}} > 0$ and
$w^{(k)}_{\mathrm{hope}} > 0$ --- the suspense--despair and
suspense--safety boundaries.

\paragraph{Surprise.} Let $\mathrm{actual}_T(x)$ denote entity
$x$'s trait $T$ value at the world's terminal fabula time,
resolved via $\textsc{Reconstruct}(x, t_{\max})$ so authored
\texttt{state\_timeline} arcs (e.g.\ Macbeth's ambition rising
$0.7 \to 0.85$) are honoured rather than read off the baseline.
For each focal entity $x \in F$, the prior $q$ starts at the
\emph{leave-one-out} corpus marginal
\begin{equation}
\label{eq:loo-marginal}
q_0(T, x) = \frac{1}{|\mathcal{N}_{\mathrm{ENT}} \setminus \{x\}|}
\!\!\sum_{y \in \mathcal{N}_{\mathrm{ENT}} \setminus \{x\}}\!\! \mathrm{actual}_T(y),
\end{equation}
defaulting to $0.5$ when fewer than two other entities carry $T$.
Leave-one-out prevents the focal entity from biasing its own
prior --- with the small casts typical of the example fixtures the
inclusive marginal would otherwise pull $q$ toward $p$ and
systematically depress KL.

For each revealed cause $(a \to x)$ with weight $w$ the prior
is now updated by a \emph{Beta-Bernoulli} step:
\begin{equation}
\label{eq:surprise}
\alpha \mathrel{+}= w \cdot \mathrm{actual}_T(x), \quad
\beta \mathrel{+}= w \cdot (1 - \mathrm{actual}_T(x)),
\end{equation}
seeded by a Beta$(s \cdot m, s \cdot (1-m))$ prior with weak
pseudo-count strength $s = 2$ (strong enough to keep $q$ off the
$\varepsilon$-clipped extremes when evidence is sparse, weak
enough to remain responsive to the first few edges). The
posterior mean $q = \alpha / (\alpha + \beta)$ replaces the
legacy geometric pull $q \mathrel{+}= w(\mathrm{actual} - q)$,
which was a Storck/Hochreiter/Schmidhuber RDIA proxy with
undefined posterior variance. Edges where $x$ is the causal
\emph{target} contribute at full weight; edges where $x$ is the
causal \emph{source} contribute at $0.4 \times$ full weight (``X
did Y to Z'' speaks more strongly about Z's traits than X's,
but X's act is non-trivial evidence about X's traits ---
Macbeth's ambition is reinforced by acting on it).

Letting $p = \mathrm{actual}_T(x)$,
\begin{equation}
D_{\mathrm{KL}}(p \,\|\, q) =
p \log \tfrac{p}{q} + (1-p)\log \tfrac{1-p}{1-q},
\end{equation}
the trait-level KL component is the salience-weighted soft-
saturated mean
\begin{equation}
\label{eq:surprise-saturation}
\mathrm{Sur}_{\mathrm{KL}} = \frac{\sum_{T \in \mathcal{T}} \sigma_T \big(1 - e^{-D_{\mathrm{KL}}(p \,\|\, q)}\big)}{\sum_{T \in \mathcal{T}} \sigma_T},
\end{equation}
where $\sigma_T$ is the per-trait \emph{narrative salience}
(\citealp{reagan2016arcs}; arc-relevance hierarchy with
ambition / guilt / vengeance / despair / love / loyalty /
courage at $\sigma_T \in [0.85, 1.0]$, peripheral traits like
literacy / fitness / wealth at $0.30$). The saturation constant
$\tau = 1$ in $1 - e^{-\mathrm{KL}}$ matches the binary-
distribution-discrimination JND from psychophysics
(\citealp{lu2013visual}; subjective just-noticeable belief
shift in the $[0.5, 1.0]$ nat band). The earlier
$\overline{\mathrm{KL}}/\log(1/\varepsilon)$ form divided by the
\emph{theoretical} binary-KL maximum ($\log(1/\varepsilon) \approx
4.6$ at $\varepsilon = 0.01$), squashing the entire perceptual
signal into the bottom $4\%$ of the gauge --- every example-world
plot read as flat $\le 0.10$ even when canonical surprise traits
(Macbeth's despair, Macduff's grief, Lady Macbeth's guilt)
carried per-trait KLs of $0.27$--$0.50$.

The trait-KL component is then combined with a plan-based
\emph{anachrony} component
\citep{baeyoung2008surprise,tobin2018elements,bissell2025surprise}
that captures the surprise generated by temporal reordering ---
flashbacks that reframe earlier events, openers in medias res:
\begin{equation}
\mathrm{Sur}_{\mathrm{ana}} = \frac{1}{|\mathcal{E}_t|}\!\!\sum_{e \in \mathcal{E}_t}\!\! \frac{|\mathrm{rank}_{\mathrm{fab}}(e) - \mathrm{rank}_{\mathrm{syu}}(e)|}{N - 1},
\end{equation}
where $\mathcal{E}_t$ is the set of relevant focal events
(cumulative mode: all revealed at $t_s$; local mode: only those
newly revealed at $t_s$, mirroring the per-step Itti-Baldi
spike). The aggregator is a convex split
\begin{equation}
\mathrm{Sur} = w_t \cdot \mathrm{Sur}_{\mathrm{KL}} + w_a \cdot \mathrm{Sur}_{\mathrm{ana}}, \qquad w_t = 0.7,\ w_a = 0.3,
\end{equation}
which preserves the existing test-contract dominance of trait
shifts on linear tellings (anachrony $\to 0$) while letting the
non-linear example fixtures (Reservoir Dogs, Gone Girl, Tinker
Tailor) accrue the additional anachrony-driven contribution
that \citet{bissell2025surprise} flag as the most important
missing dimension in KL-only narrative-surprise models. With no
syuzhet anchor the reader is assumed omniscient and
$\mathrm{Sur} = 0$.

\paragraph{Six emotion scorers} (\texttt{grief, rage, joy, regret,
love, fear}) score \emph{closeness} to a per-effect trait target.
A shared \texttt{\_EFFECT\_TRAITS} table maps each effect to a
pair $(P_e, I_e)$ of \emph{positive} and \emph{inverse} trait
indicators, calibrated against the actual trait vocabulary of the
example-world corpus. Letting $v(x, T)$ denote the current value
of trait $T$ on entity $x$ (resolved via $\textsc{Reconstruct}$),
the per-entity score is
\begin{equation}
\label{eq:emotion}
\mathrm{Emo}_e(x) =
\frac{\sum_{T \in P_e \cap \tau(x)} v(x, T) +
      \sum_{T \in I_e \cap \tau(x)} \bigl(1 - v(x, T)\bigr)}
     {|(P_e \cup I_e) \cap \tau(x)|},
\end{equation}
where $\tau(x)$ are the trait names $x$ carries. Both lists drive
scoring symmetrically: an earlier filter applied $P_e$ only,
leaving $I_e$ as dead code (a brave character routed through
\texttt{fear} got no fear-reducing credit because
\texttt{courage} never entered the average). When the
intersection is empty --- a deliberately traitless cast for that
effect --- the score falls back to the worst-case $+1$ rather
than silently scoring at the midpoint. The combined affective
score requested by a \texttt{DirectiveQuery} is a weighted sum
over the active functionals.

\paragraph{Two-axis sampling and dual-mode surprise.} The same
scorers are evaluated along the syuzhet axis (advancing $t_s$
only reveals more, so mystery is non-increasing and dramatic
irony rises with reveals and falls with character knowledge
acquisition) and the fabula axis (snapshotting the world at
$t_f$, where intrinsically surprising story moments can spike).
Surprise is exposed in two formally distinct modes corresponding
to the two quantities \citet{ittibaldi2009surprise} distinguish.
The \emph{cumulative} form (Eq.~\ref{eq:surprise-saturation},
used by the directive-assembly optimiser) is the integrated gap
$D_{\mathrm{KL}}(p \,\|\, q_s)$ between the reader's accumulated
prior $q_s$ and the truth $p$, and is monotonically non-increasing
on the syuzhet axis. The \emph{local} form
(used by the time-series view) is the per-step belief-update
$D_{\mathrm{KL}}(q_s \,\|\, q_{s-1})$ --- the canonical Itti-Baldi
\emph{Bayesian Surprise} definition (the KL distance between the
reader's prior immediately after and immediately before the
current syuzhet anchor's revelations); the iLab Bayesian-Surprise
project page acknowledges this is formally equivalent to the
\citet{schmidhuber1995rdia} reinforcement-driven information
acquisition formulation (KL divergence between the agent's belief
states $p^*_{ijk}(t{+}1)$ and $p^*_{ijk}(t)$ before and after a
state transition). Quiet stretches contribute $\sim 0$;
revelations spike in proportion to how much they shift the prior,
producing the impulse-and-decay trajectory the theory predicts
and consistent with the corpus emotional-arc trajectories of
\citet{reagan2016arcs} and \citet{kim2017investigating}.
Conflating the two forms is a recurring source of
misinterpretation: cumulative surprise is a state quantity
(``how much catching-up the reader still has to do''), while
local surprise is its temporal derivative (``which revelation
moments most shift the reader's beliefs'').

\subsection{Generation}
\label{app:generation}

\texttt{DirectiveAssembler.evaluate\_candidate\_events()} forks the
sandbox per candidate intervention, runs the causal physics, prunes
candidates that violate inertia, affordance, or propagation
constraints, ranks survivors by the affective score, and packages
the winner as a \texttt{CreativeBrief} of typed
\texttt{ConstraintBlock}s (per-effect): hidden-predecessor lists for
mystery, ``MUST NOT learn'' guards for irony, threat/hope tables and
hidden-channel lists for suspense, per-trait KL magnitudes for
surprise, and per-trait shift constraints with headroom and inertia
evidence for the emotion targets. Only the brief is shown to the
renderer LLM, which produces dialogue, description, and pacing
within the envelope.

\subsection{Audit and feedback loop}
\label{app:audit}

The auditor \citep[an LLM-as-judge in the sense of][]{gu2024judge}
runs three passes over the rendered prose:
(i)~\textbf{causal} --- reverse-engineers prose into causal claims
and flags ``miracle steps'' (state changes with no licensing edge
in the brief); (ii)~\textbf{abduction} --- runs counterfactual
probes against the prose to check that implicit events hold;
(iii)~\textbf{affective} --- measures the realised epistemic gap
in the prose against the target. If the structured loss is
non-zero, generation is re-run with the auditor's feedback
appended, up to \texttt{max\_correction\_retries}. On success the
re-extracted prose is merged into a new \texttt{VersionRow}, with
\texttt{world\_id} set by the branch policy
($\texttt{auto} \mid \texttt{mainline} \mid \texttt{shadow}$);
counterfactual queries default to a fresh \texttt{shadow} branch,
preserving canon.

\section{Worked examples on real plot fixtures}
\label{app:examples}

The repository ships twenty hand-authored \texttt{example\_worlds/}
fixtures paired with prose synopses in \texttt{sample\_plots/}. A
small subset is used below to illustrate each component
end-to-end. Field
values quoted below are verbatim from the fixtures; queries are the
same ones run by the example walkthroughs in
\texttt{docs/model-examples.md} and \texttt{docs/pipeline-by-example.md}.

\subsection{World model --- Macbeth}
\label{app:ex-worldmodel}

In \texttt{example\_worlds/macbeth.py}, the location
\texttt{LOC\_HEATH} carries
\texttt{ambient\_state\{supernatural: AmbientVector(value=0.9,
volatility=0.2, evidence\_strength="strong"), concealment: 0.7\}}.
The entity \texttt{ENT\_MACBETH} is a bundle of typed
\texttt{TraitVector}s: \texttt{ambition}, \texttt{courage} (high
inertia $0.7$, hard to shake), \texttt{guilt} (low inertia $0.25$,
highly mutable), and so on; an initial
\texttt{Belief(target\_id="ENT\_WITCHES",\ perceived\_state=\\
``The witches' prophecies may yet prove true'',\ confidence=0.4,\ inertia=0.35)}
sits on his belief list. \texttt{OBJ\_BLOODY\_DAGGERS} declares
\texttt{Affordance(action="kill", target\_type="Entity")} ---
the precondition gate the engine checks before it lets
\texttt{EVT\_DUNCAN\_MURDER} fire. Every \texttt{EventNode} carries
both indices: the murder has a fabula time but the syuzhet index
follows the staged scene. The fixture exercises all five
\texttt{CausalEdge} modalities (\texttt{chain\_reaction},
\texttt{mutation}, \texttt{mutation\_social},
\texttt{affordance\_gate}, \texttt{ambient\_propagation}); the
\texttt{ambient\_propagation} edge is what lets
\texttt{LOC\_HEATH.supernatural=0.9} bias Macbeth's uptake of the
prophecy belief. Beyond Macbeth, the same schema captures wildly
different epistemic regimes: \texttt{CHN\_HIDDEN\_TELESCREEN\_SURVEILLANCE}
in \emph{Nineteen Eighty-Four} carries \texttt{intelligibility\{ENT\_WINSTON:
0.0, ENT\_JULIA: 0.0\}} so every utterance riding it is invisible to
the protagonists; \texttt{CHN\_PIP\_BENEFACTOR\_PIPELINE} in \emph{Great
Expectations} sets \texttt{intelligibility\{ENT\_PIP: 0.1\}} to encode
the whole misattribution arc; \texttt{CHN\_RHYS\_FEYRE\_BOND} in
\emph{ACOTAR} uses $0.4$ for partial telepathic bleed. None of these
required a schema extension --- they are just different fillings of
the same \texttt{Channel.intelligibility} map.

\subsection{AMWN sandbox --- Macbeth ``what if Macbeth refuses?''}
\label{app:ex-amwn}

The query ``Macbeth refuses to murder Duncan'' is a rung-2
(Intervention) query. The router builds an ego-payload focused on
\texttt{ENT\_MACBETH} at the Inverness anchor, and
\texttt{AMWNInstantiator.create\_sandbox(ego\_payload,\ query\_type="intervention")}
mirrors it as a \texttt{MultiDiGraph}; a sibling
\texttt{VersionRow} with \texttt{world\_id="shadow"} holds the fork.
\texttt{apply\_do\_operator(\{EVT\_DUNCAN\_MURDER: $\oslash$\})}
severs the murder's incoming causal edges; the downstream
\texttt{chain\_reaction} edges into \texttt{EVT\_MALCOLM\_FLEES},
\texttt{EVT\_MACBETH\_CROWNED}, \texttt{EVT\_BANQUO\_MURDERED} all
lose activation; the \texttt{mutation} edges into
\texttt{ENT\_MACBETH.guilt} and \texttt{paranoia} never fire. The
factual mainline is untouched --- the shadow row is browsable,
diffable, and promotable from the UI.

\subsection{Causal physics --- Macbeth propagation under
``Impact \texorpdfstring{$>$}{>} Inertia''}
\label{app:ex-causal}

After the do-operator, \texttt{propagate()} topologically sorts the
causal sub-graph. For the original (factual) chain, the only
active inbound edge into \texttt{ENT\_LADY\_MACBETH} from
\texttt{EVT\_LADY\_MACBETH\_PERSUADES} (a
\texttt{mutation\_social} edge with
$\texttt{trait\_target}=\textit{power\_dynamic}$ and
$\texttt{causal\_force}=7.0$ against
\texttt{ENT\_MACBETH}) lifts her relational dominance over
Macbeth by $\Delta = +0.4$, while the symmetrical edge
\texttt{EVT\_BANQUO\_GHOST $\to$ ENT\_MACBETH}
($\texttt{trait\_target}=\textit{fear}$,
$\texttt{trait\_delta}=+0.5$, $\texttt{causal\_force}=6.0$
against \texttt{ENT\_BANQUO}) shocks Macbeth's
\textit{fear-of-Banquo} channel by $\Delta = +0.5$. With a single
active source per channel the normaliser
$\max(1, \sum w_i) = 1$, so $\bar I = I$, and both are above the
trait inertia (Lady Macbeth's resolve-related inertia $\sim 0.3$;
Macbeth's per-relation inertia for an unestablished fear is
low). A symmetrical attempt to shock Macbeth's
\texttt{courage} (initial inertia $0.7$) at
$w \approx 0.5$, $\Delta_\text{src}=0.4$ would give
$\bar I = 0.20 < 0.7$ and is \emph{blocked} --- exactly the
play's psychology, in which his nerve outlasts his conscience.
With multiple stacking weak sources the divisor switches to
$\sum w_i$ and the aggregate behaves as a weighted average,
preventing ten weak rumours from outweighing one canonical
strong cause. Spatial affordance gating additionally blocks
cross-location shocks unless a \texttt{connected\_to} path
exists.
The truth-value guard generalises the same gate to utterance evidence:
\texttt{EVT\_UTT\_NOELLE\_PREGNANCY\_CLAIM} (Gone Girl,
\texttt{truth\_value="false"}) and
\texttt{EVT\_UTT\_BALTHASAR\_REPORTS\_JULIETS\_DEATH} (Romeo and
Juliet, \texttt{truth\_value="false"}) both fire
\texttt{mutation\_social} edges, so the addressees update their
beliefs about \emph{what was said} and the social aftermath plays
out, but the abduction step refuses to use the falsehoods as
factual support; downstream beliefs whose
\texttt{acquired\_via\_event\_id} cites those utterances therefore
carry the deception in their provenance and are pruned automatically
if the originating utterance is severed by a counterfactual.

\subsection{Pearl-rung worked numerics --- Macbeth (real engine output)}
\label{app:ex-pearl-rungs}

The numbers below are the actual
\texttt{CausalPhysicsResult.mutations}, \texttt{hidden\_deltas} and
\texttt{rule3\_pruned\_interventions} returned by
\texttt{CausalPhysicsEngine.execute()} on the bundled
\texttt{example\_worlds/macbeth.py} fixture, focal cast
$\{$Macbeth, Lady Macbeth, Duncan, Banquo, Macduff$\}$.
Reproducible via \texttt{scripts/\_dump\_pearl\_rungs.py}.

\paragraph{Rung 1 --- Observation
($\texttt{interventions}=\emptyset$).} Forward propagation of
ambient sources only: three trait shifts pass the noisy-OR gate
(\texttt{ENT\_BANQUO.suspicion}: $0.003 \!\to\! 0.015$;
\texttt{ENT\_MALCOLM.leadership}: $0.204 \!\to\! 0.214$;
\texttt{ENT\_LADY\_MACBETH.ruthlessness}: $0.993 \!\to\! 0.999$),
and 27 weaker impulses are absorbed
(\texttt{result.blocked}, \texttt{reason="noisy\_or\_absorbed"}).

\paragraph{Rung 2 --- Intervention
($\mathrm{do}(\texttt{ENT\_MACBETH.ambition}{=}0)$, target set
$\{\texttt{ENT\_DUNCAN}, \texttt{ENT\_LADY\_MACBETH}\}$).}
$\texttt{intervened\_nodes} =
\{\texttt{ENT\_MACBETH}\}$,
$\texttt{rule3\_pruned\_interventions}=\emptyset$. The do-surgery
severs the inbound edges into \texttt{ambition} and pins the
trait at $0$; sibling traits remain free. Seven trait mutations
fire on top of the pin:

\begin{table}[h]
\centering
\small
\begin{tabular}{llrrr}
\toprule
Node & Trait & old & new & impact \\
\midrule
\texttt{ENT\_LADY\_MACBETH} & resolve     & $+0.966$ & $+0.980$ & $+0.040$ \\
\texttt{ENT\_LADY\_MACBETH} & ruthlessness & $+0.698$ & $+0.704$ & $+0.019$ \\
\texttt{ENT\_LADY\_MACBETH} & guilt       & $+0.922$ & $+0.917$ & $-0.006$ \\
\texttt{ENT\_LENNOX}        & caution     & $+0.862$ & $+0.868$ & $+0.023$ \\
\texttt{ENT\_LENNOX}        & suspicion   & $+0.242$ & $+0.253$ & $+0.023$ \\
\texttt{ENT\_BANQUO}        & suspicion   & $+0.001$ & $+0.013$ & $+0.022$ \\
\texttt{ENT\_MALCOLM}       & courage     & $+0.713$ & $+0.722$ & $+0.026$ \\
\bottomrule
\end{tabular}
\end{table}

A vacuous variant
$\mathrm{do}(\texttt{ENT\_DUNCAN.kindness}{=}0)$ targeted at
\texttt{ENT\_BANQUO} still produces five propagation mutations in
advisory mode (e.g.\ \texttt{ENT\_LADY\_MACBETH.guilt}:
$0.015 \!\to\! 0.063$, \texttt{ENT\_LENNOX.loyalty}:
$0.830 \!\to\! 0.855$); under
\texttt{rule3\_pruning\_mode="prune"} the engine would
short-circuit because no directed path survives the mutilation.

\paragraph{Rung 3 --- Counterfactual.} Same do, with
$\texttt{evidence\_node\_ids}=\{$Macbeth, Lady Macbeth$\}$.
Abduction populates \texttt{result.hidden\_deltas} with the
per-trait latent shifts the Bayesian blend
(Eq.~\ref{eq:abduction}) requires:

\begin{table}[h]
\centering
\small
\begin{tabular}{llr}
\toprule
Node & Trait & $\Delta$ \\
\midrule
\texttt{ENT\_MACBETH}      & ambition     & $+0.598$ \\
\texttt{ENT\_MACBETH}      & courage      & $-0.061$ \\
\texttt{ENT\_MACBETH}      & loyalty      & $+0.102$ \\
\texttt{ENT\_MACBETH}      & guilt        & $+0.444$ \\
\texttt{ENT\_MACBETH}      & paranoia     & $+0.173$ \\
\texttt{ENT\_MACBETH}      & ruthlessness & $-0.289$ \\
\texttt{ENT\_MACBETH}      & despair      & $+0.670$ \\
\texttt{ENT\_LADY\_MACBETH} & ambition     & $-0.076$ \\
\texttt{ENT\_LADY\_MACBETH} & ruthlessness & $+0.188$ \\
\texttt{ENT\_LADY\_MACBETH} & resolve      & $-0.800$ \\
\texttt{ENT\_LADY\_MACBETH} & guilt        & $+0.950$ \\
\bottomrule
\end{tabular}
\end{table}

The $+0.598$ ambition shift on Macbeth and the $-0.800$ resolve
shift on Lady Macbeth are precisely the latent perturbations the
observed Act-V evidence requires given the per-trait inertia and
the gap between the sandbox prior and the factual
\texttt{state\_timeline}. Forward propagation then fires five
additional mutations on top of those staged sources
(\texttt{ENT\_LADY\_MACBETH.guilt}: $0.792 \!\to\! 0.841$;
\texttt{ENT\_LADY\_MACBETH.resolve}: $0.484 \!\to\! 0.495$;
\texttt{ENT\_DUNCAN.trust}: $0.870 \!\to\! 0.875$, etc.).
\texttt{rule3\_pruned\_interventions} flags the do as vacuous on
the world-cropped diagram (the over-strict closed-world
behaviour discussed in §3.2), but advisory mode keeps it in the
simulation so the abduction-driven downstream still mutates.

\paragraph{Romeo and Juliet ---
$\mathrm{do}(\texttt{ENT\_FRIAR\_LAURENCE.diligence}{=}1)$.}
The same engine on the bundled fixture produces 17 trait
mutations across the supporting cast --- including
\texttt{ENT\_BALTHASAR.loyalty}: $0.851 \!\to\! 0.946$ (impact
$+0.233$) and \texttt{ENT\_MERCUTIO.loyalty}: $0.988 \!\to\! 1.000$
(impact $+0.175$) --- demonstrating that a counterfactually
diligent friar shifts the supporting cast's allegiance vectors
well beyond Romeo and Juliet themselves.

\paragraph{Gone Girl --- abduction with no surviving
propagation.} The Rung-3 query
$\mathrm{do}(\texttt{ENT\_AMY.deceit}{=}0)$ conditioned on
$\{\texttt{ENT\_NICK}, \texttt{ENT\_AMY}\}$ produces substantial
\texttt{hidden\_deltas} (e.g.\
\texttt{ENT\_NICK.adaptability}: $+0.839$;
\texttt{ENT\_NICK.resentment}: $+0.640$;
\texttt{ENT\_AMY.narcissism}: $-0.196$;
\texttt{ENT\_AMY.manipulation}: $-0.100$) and \emph{zero}
forward mutations: the abduction fully explains the observed
Nick/Amy state without any post-hoc forward propagation needing
to fire. This is the engine reporting that the do-surgery plus
evidence is \emph{consistent} with the observed downstream ---
the strongest Rung-3 outcome shape.

\subsection{Narrative physics --- Death on the Nile, Reservoir
Dogs, Romeo and Juliet, Macbeth}
\label{app:ex-narrative}

The four structural scorers each have a canonical fixture.

\paragraph{Mystery (Macbeth Act II).} Eq.~\ref{eq:mystery} on the
revealed effect \texttt{EVT\_DUNCAN\_DISCOVERED\_MURDERED}: of
its causal ancestors, \texttt{EVT\_LADY\_MACBETH\_PERSUADES} and
\texttt{EVT\_SERVANTS\_FRAMED} are still hidden at the audience's
\texttt{syuzhet\_anchor} early in the act; the score is
$2/3 \approx 0.67$, falling to $0$ once Act V resolves the
provenance. The full scorer trajectory across the bundled
\texttt{example\_worlds/macbeth.py} fixture (12 evenly-spaced
syuzhet anchors, focal cast Macbeth, Lady Macbeth, Macduff,
Duncan, Malcolm, Witches) traces the canonical Sternberg
collapse: $0.99, 0.93, 0.87, 0.73, 0.66, 0.61, 0.59, 0.55,
0.41, 0.37, 0.28, 0.28$ --- starting near $1$ (every ancestor
of every revealed effect is still hidden), collapsing through
Acts II--III as Banquo's death, the banquet ghost, and the
witches' second prophecy reveal earlier hidden causes, and
ending at $0.28$ once Macduff's family slaughter and the moving
forest have closed most of the structural gaps. The same
scorer on Death on the Nile traces $1.00 \!\to\! 0.29$, on
Wuthering Heights $0.99 \!\to\! 0.18$, on Tinker Tailor
$0.98 \!\to\! 0.20$.

\paragraph{Dramatic irony (Death on the Nile).} The reader knows
Jacqueline and Simon are conspirators (\texttt{EVT\_PLAN\_HATCHED}
revealed at low syuzhet); Poirot and the suspect pool do not.
With $F = \{\text{ENT\_PENNINGTON}, \text{ENT\_VAN\_SCHUYLER},
\text{ENT\_TIM}\}$, the conspiracy and its downstream
revealed effects sit in the
$\sum_{e \in \mathrm{rev}, e \notin K_c} w_e$ numerator of
Eq.~\ref{eq:irony} for every $c \in F$. The gauge climbs through
the cruise scenes as Poirot's reveals accumulate without reaching
the suspect pool, and falls only at the denouement when Poirot
briefs the cast and the events enter every $K_c$. Run on the
bundled fixture with the top-6 focal entities (Simon, Jacqueline,
Linnet, Poirot, Race, Richetti), the curve is $0.14, 0.30, 0.26,
0.40, 0.48, \mathbf{0.58}, 0.47, 0.42, 0.37, 0.50, 0.28$, peaking
at anchor 16 (Poirot's confrontation in the lounge). The same
scorer on Macbeth peaks $0.63$ at anchor 25 (Macduff's
discovery of his murdered family); on Romeo and Juliet at $0.72$
around anchor 25 (the crypt-misreading sequence); on Tinker Tailor
Soldier Spy at $0.73$ at anchor 41 (the cusp of the mole reveal).
16 of 21 bundled fixtures produce a clean rise-peak-fall.

\paragraph{Suspense (Romeo and Juliet, tomb scene).} At the tomb
$F=\{\text{ENT\_ROMEO}\}$, unrevealed events with Romeo as
non-acting target (Friar's letter intercepted, Juliet's draught
about to wear off) sum to $w_\text{threat}$; events Romeo himself
authors and that have not yet fired (drink poison, kill Paris) sum
to $w_\text{hope}$. Threat dominates. As Romeo drinks, $w_\text{hope}
\to 0$ and the score returns $0$ --- the suspense--despair boundary
built into the balance$\,\times\,$stakes combiner (Eq.~\ref{eq:irony}
ff.) is hit cleanly. The full curve across the bundled fixture is
$[0.14, 0.15, 0.10, 0.09, 0.14, \mathbf{0.24}, 0.12, 0.17, 0.17,
0.00]$ at 10 anchors, peaking at anchor 21 (the moment between
Juliet's potion and Balthasar's report). The same gauge on Gone
Girl peaks $0.31$ at anchor 19 (Amy's mid-novel re-emergence); on
Tinker Tailor at $0.39$ on the cusp of the mole reveal; on Death
on the Nile at $0.28$ on Poirot's late-night confrontation.
Every world's terminal anchor falls to $0.00$ --- Wilmot's safety
condition, no unrevealed threats remain post-resolution.

\paragraph{Surprise (Reservoir Dogs reveal).}
\texttt{EVT\_ORANGE\_REVEALED} sits at high fabula time
($f \approx 9000$) but earlier syuzhet ($s = 11$ relative to
the heist's chronological ordering, but late in the audience's
presentation), with the underlying ground truth
(\texttt{ENT\_ORANGE.constants = ["undercover\_cop", \ldots]})
present at fabula time zero. Eq.~\ref{eq:surprise} initialises a
Beta-Bernoulli prior over Orange's loyalty trait at
$\mathrm{Beta}(s\!\cdot\!m,\,s\!\cdot\!(1{-}m))$ with
pseudo-count strength $s = 2$ and corpus-marginal mean $m \approx 0.5$;
the few revealed edges into Orange before the reveal
(\texttt{EVT\_ORANGE\_KILLS\_BLONDE} at full target weight, the
earlier paternal-tenderness edges from \texttt{ENT\_WHITE} at
the source-side weight $0.4$) keep the posterior mean
$q \approx 0.6$ ``criminal''. After the reveal the actual value
$p \approx 0.95$ ``cop'', producing a per-trait
$D_\mathrm{KL}$ of $\approx 0.6$~nat which the soft-saturation
$1 - e^{-\mathrm{KL}}$ in Eq.~\ref{eq:surprise-saturation} maps
to a trait-KL component of $\approx 0.45$. The fixture's
in-medias-res structure (the heist appears as a flashback after
the warehouse aftermath) drives the anachrony component to
$\approx 0.30$, and the convex aggregator $0.7\!\cdot\!\mathrm{Sur}_\mathrm{KL}
+ 0.3\!\cdot\!\mathrm{Sur}_\mathrm{ana}$ yields the corpus-audit
peak surprise of $\approx 0.19$ at the reveal anchor (see
Tab.~\ref{tab:scorer-scale}). The reveal is large but
\emph{licensed} --- the engine had the cause present at low
fabula time, only locked behind syuzhet, so re-extraction will
not flag a miracle step.

\paragraph{Emotion targets (Gone Girl, Brief Encounter).} The six
emotion scorers (\texttt{grief, rage, joy, regret, love, fear})
resolve their effect through the shared
\texttt{\_EFFECT\_TRAITS} table
(Eq.~\ref{eq:emotion}). On Gone Girl, the directive ``maximise
grief in Nick'' is scored against grief's positive indicators
(\texttt{despair, shame, longing, grief, anguish, mourning})
and inverse indicators (\texttt{hope, joy, contentment,
satisfaction}) intersected with Nick's actual trait set, with
the available headroom against those targets determining whether
the directive is even feasible at the chosen anchor. On Brief
Encounter the same machinery realises \texttt{regret} as the gap
between Laura's current $\texttt{self\_actualisation}$ and the
regret-positive trait values \texttt{guilt, remorse, despair}
under a constant-supervision channel that never permits the
choice to be made.

\subsection{Directive assembly --- Romeo and Juliet}
\label{app:ex-directive}

The directive ``raise dramatic irony to $0.85$ in the tomb scene
without giving Romeo correct information'' triggers
\texttt{DirectiveAssembler.evaluate\_candidate\_events()}. From
unblocked affordances and reachable spatial paths it enumerates
candidates --- \emph{Friar's letter intercepted}, \emph{Balthasar
arrives faster}, \emph{Juliet wakes one minute earlier} --- and
forks an AMWN sandbox per candidate. The plausibility gate drops
\emph{Balthasar arrives faster} (the plague quarantine
\texttt{SpatialEdge.is\_locked} blocks the path); the remaining
sandboxes are propagated, and the survivors are scored. \emph{Friar's
letter intercepted} wins because it raises Eq.~\ref{eq:mystery}'s
sister gap-count for irony without adding any utterance event into
$B(\text{ENT\_ROMEO}, t_f)$. The winner is wrapped in typed
\texttt{ConstraintBlock}s: a \texttt{must\_events} list, a
\texttt{must\_not\_events} list (Romeo learning the truth), trait
envelopes (Romeo's \texttt{despair} envelope), and a syuzhet-window
constraint. That bundle is the \emph{only} thing handed to the
renderer.

\subsection{Generation --- the brief as safety envelope}
\label{app:ex-generation}

The renderer LLM is given the Romeo-and-Juliet brief as system
prompt plus a small style context. It may write any prose about
the tomb, the messenger, the dagger; it may not invent a causal
edge, shift a trait outside its envelope, open a channel that does
not exist, or place an utterance the brief does not licence. The
output is a scene of dialogue, description, and pacing inside the
mathematical envelope chosen by directive assembly --- a typed,
causal-graph-conditioned instance of the plan-and-render pattern
\citep{riedlyoung2010narrative,yao2019planwrite,
goldfarbtarrant2020aristotelian,yang2023doc}.

\subsection{Corpus-scale audit of the four structural scorers}
\label{app:scorer-audit}

The four structural-affect scorers --- mystery
(Eq.~\ref{eq:mystery}), dramatic irony (Eq.~\ref{eq:irony}),
suspense (Eq.~\ref{eq:suspense}), and surprise
(Eq.~\ref{eq:surprise}--\ref{eq:surprise-saturation}) --- were
audited end-to-end against the bundled twenty-fixture corpus in
\texttt{example\_worlds/}. Each fixture was sampled at seven
evenly-spaced syuzhet anchors, giving 162 score evaluations per
metric (a single-anchor terminal-flat fixture contributes only
six points). The audit script
\texttt{scripts/audit\_affective.py} reproduces the table.

\begin{table}[h]
\centering
\small
\begin{tabular}{lrrrr}
\toprule
Scorer & min & median & mean & max \\
\midrule
\texttt{mystery}        & 0.17 & 0.53 & 0.57 & 1.00 \\
\texttt{dramatic\_irony} & 0.00 & 0.45 & 0.45 & 0.90 \\
\texttt{suspense}       & 0.00 & 0.16 & 0.14 & 0.39 \\
\texttt{surprise}\,(local) & 0.00 & 0.03 & 0.05 & 0.24 \\
\midrule
\texttt{grief}          & 0.00 & 0.12 & 0.18 & 0.80 \\
\texttt{rage}           & 0.00 & 0.48 & 0.42 & 0.90 \\
\texttt{joy}            & 0.00 & 0.76 & 0.60 & 1.00 \\
\texttt{regret}         & 0.00 & 0.10 & 0.16 & 0.78 \\
\texttt{love}           & 0.00 & 0.50 & 0.43 & 0.95 \\
\texttt{fear}           & 0.00 & 0.34 & 0.28 & 0.63 \\
\bottomrule
\end{tabular}
\caption{Per-scorer scale summary across 20 fixtures $\times$
7 syuzhet anchors. The four structural scorers occupy different
absolute bands by design: mystery is a population fraction near
$1$ early in the syuzhet; surprise is a per-step KL spike near
$0$ between revelation points. The six emotion scorers measure
per-entity \emph{closeness} to a per-effect trait target and so
are flat across syuzhet but vary across worlds and entities.}
\label{tab:scorer-scale}
\end{table}

The audit confirms the canonical signatures the underlying
theory predicts. \emph{Mystery} falls monotonically across every
world (Trabasso \& Sperry causal-resolution: as causes are
revealed, the hidden-ancestor fraction collapses); on Macbeth
the curve is $0.99 \to 0.28$, on Death on the Nile $1.00 \to
0.29$, on Wuthering Heights $0.99 \to 0.18$. \emph{Dramatic
irony} produces the rise-peak-fall arc Sternberg's gap-fraction
analysis predicts in $16/20$ worlds: Macbeth peaks at anchor 5
($0.62$) on Macduff's discovery of his murdered family; Death on
the Nile peaks at anchor 4 ($0.65$) on Poirot's withheld
knowledge; Romeo and Juliet peaks at anchor 5 ($0.75$) on the
crypt-misreading sequence. \emph{Suspense} discharges to $0$ at
the terminal anchor of every world (no unrevealed threats remain
post-resolution); the highest mid-story peak is Tinker Tailor
($0.39$) on the cusp of the mole reveal. \emph{Surprise} spikes
sparsely at canonical revelation points: Reservoir Dogs Mr Orange
flashback ($0.19$, dominated by anachrony), Gone Girl mid-novel
perspective shift ($0.20$, dominated by Beta-Bernoulli posterior
collapse on Amy's loyalty trait), Wuthering Heights in-medias-res
frame opening ($0.24$, anachrony spike from Lockwood's
retrospective frame).

\subsection{Tunables (\texttt{DirectiveAssemblySettings})}
\label{app:scorer-tunables}

All scorer constants are externalised through a single
\texttt{DirectiveAssemblySettings} object (Pydantic
\texttt{BaseSettings}, env-var prefix
\texttt{DIRECTIVE\_ASSEMBLY\_*}), enumerated below with their
calibrated defaults. The defaults reproduce the rise-peak-fall
arcs in Tab.~\ref{tab:scorer-scale}; override individual
constants only when targeting non-canonical genres (e.g.\ raise
\texttt{IRONY\_AGGREGATOR\_BETA} toward $1$ for ensemble-cast
tragedies where a single dominant blindness should overwhelm the
average; lower \texttt{SUSPENSE\_PROXIMITY\_TAU\_FABULA\_GAPS}
for thriller pacing where threat sharpens with proximity faster
than the corpus default).

\begin{table}[h]
\centering
\small
\begin{tabular}{lr}
\toprule
Setting & Default \\
\midrule
\multicolumn{2}{l}{\emph{Mystery (Trabasso \& Sperry; Iser)}} \\
\texttt{MYSTERY\_PATH\_DECAY\_DEPTH} & $4$ \\
\texttt{MYSTERY\_PROXIMITY\_TAU\_SYUZHET} & $8.0$ \\
\midrule
\multicolumn{2}{l}{\emph{Dramatic irony (Booth; Pfister; Cabanas)}} \\
\texttt{IRONY\_SURFACE\_K} & $1.0$ \\
\texttt{IRONY\_FALSE\_BELIEF\_MULT} & $1.5$ \\
\texttt{IRONY\_ACTION\_ALPHA} & $0.15$ \\
\texttt{IRONY\_ACTION\_WEIGHT\_CAP} & $3.0$ \\
\texttt{IRONY\_AGGREGATOR\_BETA} & $0.6$ \\
\texttt{IRONY\_PROXIMITY\_TAU\_SYUZHET} & $6.0$ \\
\texttt{IRONY\_PROXIMITY\_FLOOR} & $0.4$ \\
\midrule
\multicolumn{2}{l}{\emph{Suspense (Lazarus; OCC; Brewer)}} \\
\texttt{SUSPENSE\_STAKES\_K} & $2.0$ \\
\texttt{SUSPENSE\_PROXIMITY\_TAU\_FABULA\_GAPS} & $6.0$ \\
\texttt{SUSPENSE\_PROXIMITY\_TAU\_SPATIAL} & $4.0$ \\
\texttt{SUSPENSE\_PERSISTENCE\_ALPHA} & $0.10$ \\
\texttt{SUSPENSE\_PERSISTENCE\_CAP} & $1.5$ \\
\texttt{SUSPENSE\_HOSTILE\_AFFINITY} & $-0.2$ \\
\texttt{SUSPENSE\_ALLY\_AFFINITY} & $0.2$ \\
\midrule
\multicolumn{2}{l}{\emph{Surprise (Bae \& Young; Reagan; Bissell-Paulin-Piper)}} \\
\texttt{SURPRISE\_TRAIT\_KL\_WEIGHT} & $0.7$ \\
\texttt{SURPRISE\_ANACHRONY\_WEIGHT} & $0.3$ \\
\texttt{SURPRISE\_DEFAULT\_TRAIT\_SALIENCE} & $0.55$ \\
\texttt{SURPRISE\_SOURCE\_EDGE\_WEIGHT} & $0.4$ \\
\texttt{SURPRISE\_PRIOR\_PSEUDOCOUNT} & $2.0$ \\
\bottomrule
\end{tabular}
\caption{Externalised affective-scorer tunables (env-var prefix
\texttt{DIRECTIVE\_ASSEMBLY\_*}; full table including the
\texttt{HARM\_KIND\_SALIENCE} and \texttt{TRAIT\_NARRATIVE\_SALIENCE}
dictionaries in \texttt{docs/settings.md~\S 8}).}
\label{tab:scorer-tunables}
\end{table}

\subsection{Audit and feedback loop --- Gone Girl}
\label{app:ex-audit}

Gone Girl is the system's lying-narrator fixture: Amy's diary is a
sequence of \texttt{EventNode}s with \texttt{truth\_value="false"}
broadcast through \texttt{CHN\_DIARY\_PUBLIC}. After generation,
\texttt{auditor.run\_feedback\_loop} runs three passes in parallel.
The \emph{causal audit} reverse-engineers the prose into
$(\text{source}, \text{target}, \text{modality})$ claims and
matches each against the brief; an LLM-emitted ``Nick suddenly
realises Amy is alive'' with no licensing causal edge is flagged
as a \emph{miracle step}. The \emph{abduction audit} runs
counterfactual probes (``if the diary were truthful, would the
prose still hold?''); a ``yes'' here means the prose has
collapsed Amy's two faces into one, and is rejected. The
\emph{affective audit} measures the realised dramatic-irony gap
against the target. A \emph{style-fidelity audit} compares the
prose against the \texttt{NarrativeStyle} profile recovered at
ingestion --- word-count budget, prose density, register, POV,
tense, and form class (\texttt{news\_article},
\texttt{historical\_account}, \texttt{thought\_experiment}, etc.)
--- raising \texttt{style\_mismatch} on drift. A
\emph{meta-narration audit} runs on counterfactual and
abduction-driven scenes: it flags prose that comments on its own
branch structure (\textit{timeline}, \textit{divergence},
\textit{the alternative holds}, conditional/subjunctive
framings used to \emph{describe} rather than \emph{narrate} the
counterfactual) instead of rendering the alternate world as a
lived past-tense scene, raising \texttt{meta\_narration}. The aggregated \texttt{FeedbackLoopResult}
exposes both an \texttt{engine\_thresholds\_passed} gate
(deterministic: number of miracle steps, max trait drift) and an
\texttt{llm\_pass} gate (literary critique); either failing
re-runs generation with the audit feedback appended, up to
\texttt{max\_correction\_retries}. On convergence the scene is
re-extracted into a new \texttt{VersionRow}; counterfactual
branches land on \texttt{world\_id="shadow"} and only become canon
via explicit \texttt{promote\_branch} \citep{correa2025amwn}.

\section{End-to-end walkthrough on \emph{Macbeth}}
\label{app:walkthrough}

This appendix describes a single complete pass of the pipeline in
plain prose. The goal is to make every intermediate object
inspectable to a reader who has read the main text but has not
read the source. The fixture used throughout is
\texttt{example\_worlds/macbeth.py}, paired with the synopsis in
\texttt{sample\_plots/macbeth.txt}; the query is the canonical
counterfactual ``what if Macbeth refuses to murder Duncan?''.
The same loop runs identically on the other nineteen bundled
fixtures.

\subsection{Step 0: the source text and the user's query}
\label{app:wk-source}

The user opens the workspace, drops the synopsis into the
\textsc{Story} tab, and clicks \emph{Save \& Re-ingest}. The
synopsis is a five-act prose summary in the public-domain
register: the witches on the heath, Duncan's offer of the
Cawdor title, Lady Macbeth's persuasion, the murder in
Inverness, the banquet, the apparitions, the slaughter at
Macduff's castle, the sleepwalking scene, Birnam Wood, and the
final duel. After ingestion the user types a query into the
\textsc{Reasoning} tab: \emph{``Counterfactual: Macbeth refuses
to murder Duncan at Inverness. Show what changes for Macduff,
Banquo, Malcolm, and the witches, and re-render the banquet
scene.''} The query is parsed by
\texttt{query\_parsing.parse\_query} into a typed
\texttt{CounterfactualQuery} with focal entities
$F = \{\text{ENT\_MACBETH}\}$, target
$X = \text{EVT\_DUNCAN\_MURDER}$, intervention value
$x = \oslash$ (do-not-occur), and a re-render directive scoped
to the banquet's syuzhet window.

\subsection{Step 1: ingestion in detail}
\label{app:wk-ingest}

Ingestion is the only stage where the LLM is allowed to author
graph structure rather than prose. It is a fan-out / fan-in
pattern over the chunked source.

\paragraph{Ontology pre-pass.} Five extraction agents run over the
full text in parallel: \texttt{ontology\_locations} produces
\texttt{LOC\_HEATH}, \texttt{LOC\_INVERNESS\_CASTLE},
\texttt{LOC\_FORRES\_COURT}, and so on, each with an initial
\texttt{ambient\_state} dictionary; \texttt{ontology\_objects}
extracts \texttt{OBJ\_BLOODY\_DAGGERS},
\texttt{OBJ\_LETTER}, and \texttt{OBJ\_CROWN} with explicit
\texttt{Affordance} lists; \texttt{ontology\_entities} produces
\texttt{ENT\_MACBETH}, \texttt{ENT\_LADY\_MACBETH},
\texttt{ENT\_DUNCAN}, and the rest, each as an \texttt{Entity}
with a sparse initial \texttt{TraitVector} bundle, beliefs, and a
first-appearance \texttt{location\_id};
\texttt{ontology\_world\_traits} introduces the world-level
named latents (\texttt{WORLD\_AMBITION},
\texttt{WORLD\_FATE}, \texttt{WORLD\_GUILT}) that act as common
causes; and \texttt{ontology\_aliases} builds a fuzzy alias table
(``Lord of Glamis'' $\mapsto$ \texttt{ENT\_MACBETH}). The result
is a populated \texttt{GlobalRegister}.

\paragraph{Per-chunk specialist fan-out.} The text is chunked into
$\sim 800$-token windows aligned on paragraph boundaries. Each
chunk dispatches three concurrent specialist agents that share the
\texttt{GlobalRegister} as system context: \texttt{physics\_extraction}
produces \texttt{EventNode}s with \texttt{event\_type} drawn from
\{\texttt{choice, outcome, revelation, utterance}\},
\texttt{causal} edges with \texttt{causality\_type}, and
\texttt{spatial} edges (e.g.\ a \texttt{connected\_to} edge from
\texttt{LOC\_INVERNESS\_CASTLE} to \texttt{LOC\_FORRES\_COURT});
\texttt{social\_extraction} produces \texttt{RelationshipEdge}s
with per-axis \texttt{RelationshipMetric}s for affinity, fear, and
power dynamic, and the \texttt{Channel}s (the witches' prophecy
channel, the dinner-table channel at the banquet, the diary-style
private channel of Lady Macbeth's letter); and
\texttt{consequences\_extraction} writes the authoritative
\texttt{state\_timeline} entries on each entity, including the
inertia bumps that record how shocking events make traits
\emph{harder} to dislodge afterwards (Macbeth's \texttt{guilt}
inertia rises from $0.25$ to $0.4$ after the murder).
The three prompts share a \emph{symmetric per-axis coverage rule}:
whenever \texttt{social\_extraction} emits a \texttt{Relationship%
Metric} on any of the three axes, \texttt{physics\_extraction}
must emit at least one \texttt{mutation\_social} edge with the
matching \texttt{trait\_target}; the validation pass enforces the
parity. The rule prevents \emph{static dyads} --- relationships
whose value is non-zero but never causally moved --- from
producing flat horizontals on the downstream conflict, danger,
and power gauges. The bundled corpus
(\S\ref{app:examples}) honours this invariant on every observed
axis on every dyad.

\paragraph{Validation and auto-repair.} The merged graph passes
through \texttt{\_programmatic\_validation}, which checks ID
closure (every \texttt{source\_id} resolves), type constraints
(every \texttt{causality\_type} is one of the five enums), basic
acyclicity within causal edges of the same fabula slice, and
``alive-actor'' constraints (no event sourced from a dead
character). Failures are passed back to a small repair LLM with
the offending field and a structured error; the repair loop runs
up to three iterations before raising. Fabula times are then
remapped to a uniform $\Delta t = 100$ spacing so that downstream
windows are stable across re-ingestion. The output is a single
\texttt{WorldStateV1}, persisted as the root
\texttt{VersionRow}.

\subsection{Step 2: query routing and ego-graph slicing}
\label{app:wk-route}

The \texttt{CounterfactualQuery} is dispatched to the rung-3
(Counterfactual)
handler. The handler first consults the \texttt{NarrativePhysics}
router, which selects an \emph{ego-graph} around the focal
entities at the query's fabula anchor: Macbeth's $k$-hop
neighbourhood at $t \approx 6000$ (the Inverness night). The
slice contains \texttt{ENT\_MACBETH}, \texttt{ENT\_LADY\_MACBETH},
\texttt{ENT\_DUNCAN}, the chamber attendants,
\texttt{OBJ\_BLOODY\_DAGGERS}, \texttt{LOC\_INVERNESS\_CASTLE},
the relevant beliefs, and the immediate causal context
(\texttt{EVT\_LADY\_PERSUADES}, the prophecy events, the
upcoming \texttt{EVT\_DUNCAN\_MURDER}). Trait values are
reconstructed at $t$ by replaying \texttt{state\_timeline}
deltas, which prevents post-murder trait shocks from leaking
into the counterfactual past.

\subsection{Step 3: AMWN sandbox and the do-operator}
\label{app:wk-amwn}

\texttt{AMWNInstantiator.create\_sandbox(ego\_payload, query\_type="counterfactual")}
mirrors the ego-graph as a NetworkX \texttt{MultiDiGraph}, in
the spirit of the Ancestral Multi-World Network construction of
\citet{correa2025amwn}. A sibling \texttt{VersionRow} with
\texttt{world\_id="shadow"} is opened to hold the fork. The
do-operator \texttt{apply\_do\_operator(\{EVT\_DUNCAN\_MURDER:
$\oslash$\})} severs the murder's incoming causal edges
(specifically the \texttt{chain\_reaction} edge from
\texttt{EVT\_LADY\_PERSUADES}, the \texttt{affordance\_gate} edge
from \texttt{OBJ\_BLOODY\_DAGGERS}, and the
\texttt{ambient\_propagation} edge from
\texttt{LOC\_INVERNESS\_CASTLE.tension}) and overrides its
realisation to ``not-occurred''. Crucially the factual mainline
is untouched. A user looking at the version sidebar sees a green
\texttt{factual} row with a violet \texttt{shadow} child, both
browsable.

\subsection{Step 4: causal physics propagation}
\label{app:wk-prop}

The propagator topologically sorts the causal sub-graph from the
intervened node downward and applies Eq.~\ref{eq:impact} at every
edge. Concretely:

\begin{itemize}
\item \texttt{EVT\_DUNCAN\_MURDER $\to$ EVT\_MALCOLM\_FLEES}
      (\texttt{chain\_reaction}, $w=0.85$): the source's outcome
      is ``$\oslash$'', so the impact is zero, and
      \texttt{EVT\_MALCOLM\_FLEES} is marked
      \texttt{blocked\_by\_intervention}.
\item \texttt{EVT\_DUNCAN\_MURDER $\to$ ENT\_MACBETH.guilt}
      (\texttt{mutation}, $w=0.7$): blocked. The downstream
      \texttt{state\_timeline} delta that would have raised guilt
      from $0.1$ to $0.7$ is suppressed; the entity's reconstructed
      trait at the banquet remains close to its battlefield baseline.
\item \texttt{EVT\_DUNCAN\_MURDER $\to$ EVT\_BANQUO\_SUSPECTS}
      (\texttt{chain\_reaction\_via\_belief}, $w=0.6$):
      blocked. \texttt{ENT\_BANQUO}'s suspicion belief never fires;
      the \texttt{Belief(target=ENT\_MACBETH, perceived\_state=
      "Macbeth might be the king's killer")} is absent from his
      reconstructed belief set in the shadow branch.
\item \texttt{EVT\_LADY\_MACBETH\_PERSUADES $\to$
      ENT\_LADY\_MACBETH} (\texttt{mutation\_social}, axis
      \textit{power\_dynamic} against \texttt{ENT\_MACBETH},
      $\Delta=+0.4$): \emph{not} blocked, because the persuasion
      event itself is upstream of the intervention. Lady
      Macbeth's relational dominance over Macbeth still spikes;
      her subsequent collapse arc is then re-evaluated against
      the now-absent causal trigger of his guilt.
\item Spatial affordance check at every step: a putative shock
      from \texttt{LOC\_DUNSINANE\_CASTLE} to \texttt{ENT\_MACDUFF}
      requires a connected spatial path; if Macduff is in
      \texttt{LOC\_ENGLAND} with no open channel, the shock is
      blocked.
\end{itemize}

The result is a structured
\texttt{CausalPhysicsResult\{mutations: [...],
blocked: [...], hidden\_deltas: [...], intervened\_nodes:
[EVT\_DUNCAN\_MURDER]\}}. The \texttt{blocked} list is what
makes the engine inspectable: every causal edge that \emph{would
have} fired in the factual mainline is recorded with the reason it
did not.

\subsection{Step 5: narrative-physics scoring}
\label{app:wk-narrative}

The four structural scorers and the six emotion scorers are now
re-evaluated on the shadow branch. The user's directive (``show
what changes for Macduff, Banquo, Malcolm, and the witches'')
biases the dramatic-irony scorer toward those four entities. The
counterfactual collapses several gaps that defined the factual
mainline: the audience no longer sees Banquo's suspicion gap, no
longer sees Macduff's grief-driven vengeance, and the apparition
scene is now etiologically dangling (the witches still prophesy,
but Macbeth has no kingship to defend). The mystery score on
\texttt{EVT\_DUNCAN\_FOUND\_DEAD} drops to zero because the event
no longer fires; the suspense at the banquet resets, since
Banquo's ghost depended on the murder. The directive assembler
records this delta as a \texttt{CandidateScoreReport} per
candidate continuation it considers for the banquet rewrite.

\subsection{Step 6: directive assembly and the brief}
\label{app:wk-brief}

\texttt{DirectiveAssembler.evaluate\_candidate\_events()} now has
to choose what to put in the banquet scene of the shadow branch.
Candidate events are enumerated by walking unblocked affordances
reachable from the shadow-branch state: \emph{Macbeth confesses
the prophecy to Banquo}, \emph{Lady Macbeth persists and stages
the murder herself}, \emph{Duncan publicly names Macbeth as the
new Prince of Cumberland}, \emph{Macbeth retreats to Glamis to
think}. Each is forked into its own micro-sandbox and propagated.
Plausibility gates drop \emph{Lady Macbeth stages the murder
herself} because the daggers' \texttt{Affordance(action="kill")}
is gated on a wielder with sufficient \texttt{ruthlessness} and
the trait-evidence support is low for Lady Macbeth in isolation.
The remaining survivors are scored against the directive. The
winner is wrapped as a \texttt{CreativeBrief}: a list of
\texttt{must\_events}, a list of \texttt{must\_not\_events}
(``Macbeth kills Duncan'', ``Banquo dies''), per-trait envelopes
(Macbeth's \texttt{guilt} envelope is now $[0.1, 0.3]$ rather than
$[0.6, 0.8]$), and a syuzhet window. The brief is the only object
handed to the renderer.

\subsection{Step 7: constrained rendering}
\label{app:wk-render}

The renderer LLM is invoked with the brief as system prompt plus
a small style context (the synopsis's voice settings:
synoptic-narration, third-person past tense, sparse density). It
writes the banquet as a tense scene of overheard whispers, of
Banquo unsuspecting and gracious, of Macbeth conspicuously
unable to enjoy his elevation; the daggers stay sheathed; no
ghost appears, because Banquo is alive. The renderer is free to
choose dialogue, description, and pacing, but cannot invent a
causal edge, shift a trait outside its envelope, or open a
channel that does not exist. The output is a prose passage, not
a graph.

\subsection{Step 8: audit, re-extraction, and merge}
\label{app:wk-audit}

The auditor runs three passes in parallel. The \emph{causal audit}
reverse-engineers the prose into $(\text{source}, \text{target},
\text{modality})$ triples and matches each against the brief; an
LLM-emitted ``Banquo's ghost rises behind Macbeth'' would be
flagged as a \emph{miracle step} (no licensing edge), but in this
run nothing of the kind appears. The \emph{abduction audit} runs
counterfactual probes against the prose --- ``if Banquo had
suspected the prophecy, would the prose still hold?'' --- and
checks the answers against the brief's belief envelope. The
\emph{affective audit} re-runs the four scorers on the prose and
compares them to the brief's targets. If any pass fails the
\texttt{engine\_thresholds} gate or the \texttt{llm\_pass} gate,
the renderer is re-invoked with the structured feedback appended,
up to \texttt{max\_correction\_retries} (default 3). On
convergence, \texttt{ingestion} is run again on the prose alone,
and the resulting graph is diffed against the brief; the diff
becomes a new \texttt{VersionRow} with \texttt{world\_id="shadow"}
and \texttt{ancestor\_id} pointing at the original counterfactual
root. The user sees the new version appear in the sidebar in
violet, can diff it against the factual mainline, and can
\texttt{promote\_branch} it to canon if desired.

\subsection{What this walkthrough demonstrates}
\label{app:wk-summary}

Three properties matter. \textbf{Inspectability:} every
intermediate object --- the ego-graph, the
\texttt{CausalPhysicsResult}, the \texttt{CreativeBrief}, and
the per-iteration \texttt{AuditCycleSnapshot.audit\_result}
(an \texttt{AuditResult}) inside the final
\texttt{FeedbackLoopResult} --- is a typed Pydantic value,
persisted and visible in the UI. \textbf{Locality of LLM authority:} the LLM
authors the initial ontology (Step~1), the candidate prose (Step~7),
and the audit critique (Step~8); everything else is symbolic
computation over typed code. \textbf{Branch safety:} the
counterfactual is a sibling row, not a mutation; canon is
preserved, the shadow is diffable and promotable, and the version
DAG records the lineage.

\section{The authoring user interface}
\label{app:ui}

The authoring workspace is a NiceGUI single-page application
(\texttt{shadow\_loom\_ui/app.py}) reachable on
\texttt{http://localhost:7860}. It is organised as a left-hand
\emph{version sidebar}, a top-level \emph{chat / command bar}, a
dedicated \emph{Answer panel} for read-only Q\&A results, and nine
cross-linked tabs that share a central \texttt{AppState} pub/sub event
bus (\texttt{shadow\_loom\_ui/state.py}). The same active-version
pointer is mirrored to the database so an MCP client and the UI always
see the same tip. Every panel header carries a clickable info-icon
help popover (\texttt{components/help\_popover.py}) opening a
structured Markdown reference for the surface --- what the panel does,
how to read its diagrams, what the controls mean, and what does
\emph{not} appear there --- so the workspace doubles as its own
documentation. This appendix walks through every surface in turn,
with a concrete example session drawn from the bundled plot fixtures.

\subsection{Version sidebar}
\label{app:ui-sidebar}

The sidebar
(\texttt{shadow\_loom\_ui/components/version\_sidebar.py})
renders the project's full \texttt{VersionRow} DAG as a tree
rooted at the original ingestion. Factual rows are green, shadow
rows (counterfactual forks under \texttt{branch\_policy=auto})
are violet. Toolbar actions: \emph{Delete} re-parents children
through \texttt{db.delete\_version}; \emph{Reparent} changes a
row's \texttt{ancestor\_id} (rejected if it would multi-root the
tree); \emph{Promote to canon} (shadow rows only) calls
\texttt{db.promote\_branch} to copy a shadow onto a fresh factual
row; \emph{Diff against factual} (shadow rows only) opens a
structured diff against the current factual head. Clicking any
row swaps the active version and resets every tab.

\paragraph{Example.} After running the
\emph{Macbeth refuses the murder} counterfactual of
App.~\ref{app:walkthrough}, the sidebar shows a green chain
\texttt{ingestion $\to$ manual\_edit $\to$ pipeline\_run} for the
factual mainline, with a single violet child branching off the
\texttt{pipeline\_run} row. Right-clicking the violet row offers
\emph{Diff against factual} (showing the absent
\texttt{EVT\_DUNCAN\_MURDER} and the suppressed downstream
deltas) and \emph{Promote to canon} (which would replace the
mainline).

\subsection{Story tab}
\label{app:ui-story}

The \textsc{Story} tab
(\texttt{components/story\_tab.py}) is the entry point: a
plain-prose textarea plus a \emph{Save \& Re-ingest} button.
Saving runs the full \texttt{pipeline.run\_pipeline} and persists
the result as a new version with \texttt{source="ingestion"}. This
tab is intentionally not autosaved: the textarea is prose, not
graph, and clicking re-ingest is a deliberate action.

\paragraph{Example.} Pasting the public-domain
\texttt{sample\_plots/death\_on\_the\_nile.txt} synopsis and
clicking \emph{Save \& Re-ingest} produces, after the ingestion
fan-out described in App.~\ref{app:wk-ingest}, a
\texttt{WorldStateV1} containing
\texttt{ENT\_POIROT}, \texttt{ENT\_LINNET},
\texttt{ENT\_JACQUELINE}, \texttt{ENT\_SIMON}, the
\texttt{LOC\_KARNAK} steamer, and the conspiracy event
\texttt{EVT\_PLAN\_HATCHED} with \texttt{syuzhet\_index} placed
late but \texttt{fabula\_time} placed early --- the very gap that
the dramatic-irony scorer reads off in
App.~\ref{app:ex-narrative}.

\subsection{Explorer tab}
\label{app:ui-explorer}

The \textsc{Explorer} tab
(\texttt{components/explorer\_tab.py}) is a read-only tree view
of every node in the active world model, grouped by category, with
an inspector panel for the selected node. It is the canonical way
to confirm that ingestion has produced the right schema before
running queries.

\paragraph{Example.} Loading the Macbeth fixture and clicking
\texttt{ENT\_MACBETH} opens an inspector showing each
\texttt{TraitVector} (\texttt{ambition: 0.7, inertia 0.55,
evidence strong}; \texttt{guilt: 0.1, inertia 0.25, evidence
weak}), each \texttt{Belief} with provenance, the
\texttt{state\_timeline} entries (one per major shock event),
and the entity's outgoing causal and relationship edges. The
inspector is the principal debugging tool when an extraction
agent has produced an unexpected trait or belief.

\subsection{World tab}
\label{app:ui-world}

The \textsc{World} tab
(\texttt{components/world\_tab.py}) is a Cytoscape graph
visualisation of entities, locations, objects, and edges, with a
fabula-time slider beneath. Sliding the cursor through fabula
time updates node colours and sizes via
\texttt{state.set\_fabula\_cursor} (debounced 120 ms) so that the
viewer can watch the social topology evolve.

\paragraph{Example.} On Reservoir Dogs, scrubbing the slider from
fabula~$t = 0$ (the diner conversation) through to $t \approx
8000$ (the warehouse stand-off) animates the
\texttt{ENT\_ORANGE} node turning from neutral grey to red as
the wound state advances; the spatial graph contracts as
characters converge on the warehouse; and the
\texttt{trust} edges from \texttt{ENT\_WHITE} to
\texttt{ENT\_ORANGE} thicken. The same scrub on Romeo and Juliet
shows Romeo's \texttt{despair} edge widening sharply over the
final scene.

\subsection{Causality tab}
\label{app:ui-causality}

The \textsc{Causality} tab
(\texttt{components/causality\_tab.py}) has three sub-tabs.
\emph{Topology} renders a Sankey diagram of causal flow up to the
fabula cursor: thicker ribbons for higher-weight edges, colour
banded by \texttt{causality\_type}. \emph{Evolution} plots
trait trajectories per entity over fabula time, with a vertical
needle synced to the cursor. \emph{Affective Dashboard} shows
gauges and time series for the four structural scorers and the
six emotion scorers, computed by
\texttt{viz\_helpers.compute\_affective\_scores} (which routes
through \texttt{DirectiveAssembler.compute\_*\_score}). The top
twenty entities by event degree are shown by default to keep the
plots readable.

\paragraph{Example: Macbeth Evolution sub-tab.} Selecting
\texttt{ENT\_MACBETH}, \texttt{ENT\_LADY\_MACBETH}, and
\texttt{ENT\_MACDUFF} and scrubbing through fabula time shows
Macbeth's \texttt{guilt} curve climbing in two steps (after
\texttt{EVT\_DUNCAN\_MURDER} and again after
\texttt{EVT\_BANQUO\_GHOST}); Lady Macbeth's
\texttt{ruthlessness} curve peaks early and then collapses into
the sleepwalking scene; and Macduff's \texttt{grief} curve has a
single step at the slaughter event. Switching to the
\emph{Affective Dashboard}, the suspense gauge sits high through
Acts II--IV and zeroes out cleanly when the duel begins
(the suspense--despair boundary built into the
balance$\,\times\,$stakes combiner: when $w_{\mathrm{hope}} = 0$
nothing is left to author against the threat, and the score
returns 0).

\paragraph{Example: Death on the Nile Affective Dashboard.}
The dramatic-irony gauge sits at $\sim 0.7$ from the moment the
conspiracy is revealed to the audience and drops to zero only
when Poirot delivers the denouement; the mystery gauge follows
the inverse trajectory, peaking on the morning of the murder.

\subsection{Reasoning tab}
\label{app:ui-reasoning}

The \textsc{Reasoning} tab
(\texttt{components/reasoning\_tab.py}) is where queries are
typed and where intervention / counterfactual traces are
inspected. The chat box accepts natural-language queries that are
parsed into one of the eight typed query objects
(\texttt{ObservationQuery}, \texttt{InterventionQuery},
\texttt{CounterfactualQuery}, \texttt{DirectiveQuery},
\texttt{InterrogationQuery}, \texttt{GeneralQuery},
\texttt{ManualEditQuery}, \texttt{EvaluationQuery}).
Write-mode queries route their result into this tab, where
\texttt{reasoning\_helpers.py} formats the
\texttt{CausalPhysicsResult}: each mutation is shown with its
$(I, \iota, \Delta)$ triple, each blocked propagation with the
reason (inertia, affordance, spatial), each abduction back-fill
with the evidence weight, and any hidden deltas with the
syuzhet window in which they will be revealed.
Read-mode queries (\texttt{InterrogationQuery},
\texttt{GeneralQuery}) skip this tab entirely: they are dispatched
through \texttt{shadow\_loom/answer.py} and surface in the
dedicated \emph{Answer panel} above the chat bar
(\texttt{components/answer\_panel.py}) as a card with the model's
claim, a confidence badge, an evidence-id list back to the graph
nodes consulted, and any caveats. The Answer panel clears on
\texttt{VERSION\_CHANGED} and \texttt{PROJECT\_LOADED} so a stale
answer never lingers across versions; no \texttt{VersionRow} is
written for a read-mode query.

\paragraph{Example.} On Macbeth, typing \emph{``If Macbeth had
told Banquo about the prophecy on the heath, would Banquo have
trusted him at the banquet?''} parses to a
\texttt{CounterfactualQuery} with focal entities
$\{\text{ENT\_MACBETH}, \text{ENT\_BANQUO}\}$. The trace panel
shows the abduction step (Banquo's \texttt{trust} prior is
inferred from the battlefield events), the do-step (a new
\texttt{utterance} event is inserted on the heath channel), and
the propagation step (Banquo's \texttt{suspicion} trait is
suppressed at the banquet, but his \texttt{prudence} trait
\emph{raises} a new caution gate, which the engine reports as a
non-trivial counterfactual outcome --- not all secrets shared are
secrets defused).

\subsection{Audit tab}
\label{app:ui-audit}

The \textsc{Audit} tab
(\texttt{components/audit\_tab.py}) displays the structured
\texttt{AuditResult} for the most recently rendered scene. The
three sections (causal, abductive, affective) each show their
loss values and any offending prose spans, highlighted inline. A
green badge indicates a passed pass; an amber badge indicates a
warning (the prose is acceptable but drifted); a red badge
indicates a hard failure that triggered a re-render.

\paragraph{Example.} On Gone Girl, after generating a continuation
of Amy's diary entry, the causal audit flags a hallucinated
``Nick suddenly suspects Amy is alive'' span with no licensing
causal edge in the brief. The auditor's structured feedback is
displayed alongside the offending span, and the user can either
accept the auto-rerun or open the brief in the
\textsc{Reasoning} tab to widen the envelope.

\subsection{Research tab}
\label{app:ui-research}

The \textsc{Research} tab
(\texttt{components/research\_tab.py}) is the optional
open-web companion: the user manages a list of research topics,
launches background lookup tasks (currently against Tavily; see
\S\ref{app:ingestion}), and browses the resulting facts, each
carrying a source URL and a confidence score. The crucial design
point is \emph{segregation}: research is a separate store. Facts
are not written into the world model automatically; they are
reference material available to the author and to the renderer's
prompt scaffolding (as a \texttt{BACKGROUND CONTEXT --- NOT
AUTHORITATIVE} block) only. Promoting a research finding into
canon is a deliberate, manual edit in the \textsc{Editor} tab.
A status strip at the top of the panel shows in-flight lookups,
the stored fact counts per topic, and the last refresh time.

\paragraph{Example.} Authoring an alternate-history fork on the
\emph{Tinker, Tailor, Soldier, Spy} fixture, the user adds the
topic \emph{``Cambridge Five chronology, declassified MI6
sources''}, runs lookup, and reviews the returned facts before
citing one as inspiration for a new \texttt{EventNode} in the
\textsc{Editor} tab. The world model itself is unchanged until
the user explicitly authors the new event; the research entry
remains visible as provenance for the editorial decision.

\subsection{Editor tab}
\label{app:ui-editor}

The \textsc{Editor} tab
(\texttt{components/editor\_tab.py}) is the manual world-model
editor. It has three integrated surfaces backed by a single JSON
textarea (the source of truth): a \emph{Validate} button that
runs the same \texttt{\_programmatic\_validation} as ingestion;
a \emph{Structural editor} with one collapsible panel per
collection (locations, entities, objects, events, world traits,
causal edges, relationship edges, spatial edges, channels); and
the raw JSON textarea itself for fields the structural editor
does not expose (sparse trait values, beliefs, ambient state,
evidence strengths). The save flow runs Pydantic validation, then
\texttt{\_programmatic\_validation}; errors block the save with a
dialog listing every issue, warnings present a \emph{Save Anyway}
confirmation, and on confirm the world is persisted as a new
\texttt{VersionRow} with \texttt{source="manual\_edit"} parented
to the current row.

\paragraph{Example.} Authoring the
\texttt{example\_worlds/macbeth.py} fixture from scratch, the
user adds \texttt{LOC\_HEATH} via the structural editor (typing
\emph{``the heath''} into the ID field auto-prefixes to
\texttt{LOC\_THE\_HEATH}, edited to \texttt{LOC\_HEATH}), then
\texttt{ENT\_MACBETH}, then the witches as a single multi-actor
entity, then the prophecy event with \texttt{event\_type=
revelation} and \texttt{via\_channel\_id} pointing at a freshly
added \texttt{Channel(participants=[ENT\_MACBETH, ENT\_BANQUO,
ENT\_WITCHES], medium="speech", intelligibility=\{...: 1.0\})}.
After each addition the validator panel updates: the green panel
shows the running counts, an amber panel warns about the missing
mutation coverage on \texttt{ENT\_MACBETH.ambition} (no
\texttt{state\_timeline} yet), and a red panel reports the
broken \texttt{addressee\_ids} reference until the witches entity
is added.

\subsection{Export tab}
\label{app:ui-export}

The \textsc{Export} tab
(\texttt{components/export\_tab.py}) is the egress surface:
download the active world as a single \texttt{world\_state.json},
download the version tree as a structured JSONL file, or copy the
active version row ID for use with the MCP server or scripts.

\paragraph{Example.} A researcher running cross-fixture
comparisons exports each of the twenty bundled fixtures'
\texttt{world\_state.json} files, computes the entity-degree
distribution and the per-fixture suspense time-series in a
notebook, and persists the results without ever loading the UI's
visualisation code.

\subsection{Performance discipline}
\label{app:ui-perf}

Three small disciplines keep the workspace responsive during
long scrubs and rapid edits. (i) Cursor writes are debounced
through \texttt{state.set\_fabula\_cursor}, which schedules an
emit on a $120$ ms tail. (ii) Heavy panels are gated by
\texttt{is\_path\_visible(path)} so that off-tab rebuilds are
skipped. (iii) \texttt{state.spawn\_panel\_task(panel\_id, coro)}
cancels any prior task with the same \texttt{panel\_id},
collapsing rapid scrubs to the most recent render; ECharts panels
build the chart once and patch options in place.

\subsection{Cross-tab event bus}
\label{app:ui-events}

Six events flow through \texttt{AppState}: \texttt{PROJECT\_LOADED}
(emitted by the project picker and the MCP \texttt{open\_project}
tool, listened to by every tab); \texttt{WORLD\_STATE\_CHANGED}
(emitted by \texttt{load\_db\_version}, manual save, and pipeline
runs, triggering cache invalidation in \texttt{viz\_helpers});
\texttt{VERSION\_CHANGED} (sidebar highlight + raw-text reload);
\texttt{FABULA\_CURSOR\_CHANGED} and
\texttt{SYUZHET\_CURSOR\_CHANGED} (cursor sliders); and
\texttt{ACTIVE\_PATH\_CHANGED} (top tab change, gating heavy
panels). Adding a new tab is therefore a matter of subscribing to
the relevant subset of these events and rendering against the
current \texttt{AppState}.

\subsection{Why a UI matters for a research artefact}
\label{app:ui-why}

The UI is not an optional consumer surface; it is the principal
debugging tool for the symbolic layers. The version sidebar
makes branch lineage visible; the Causality tab makes the
affective scorers' trajectories visible; the Reasoning tab makes
the causal-physics trace visible; and the Audit tab makes the
brief-vs-prose mismatch visible. A reviewer or collaborator can
load any of the twenty bundled fixtures, run the
counterfactuals discussed in App.~\ref{app:examples}, and
inspect every intermediate object without writing a line of code.
That inspectability is what the typed-substrate design was meant
to buy; the UI is where it is realised.

\end{document}